%% file: main.tex
\documentclass[10pt,twocolumn,letterpaper]{article}

\usepackage{cvpr}      

\input{preamble}

\definecolor{cvprblue}{rgb}{0.21,0.49,0.74}
\usepackage[pagebackref,breaklinks,colorlinks,allcolors=cvprblue]{hyperref}

\def\confName{CVPR}
\def\confYear{2027}

\title{IDEA: Intent-Dominance Evidence Audit for Inference-Time Backdoor Detection in GUI Agents}

\author{Zhizhuo Kou$^{1,3}$, Sizai Hou$^{1}$, Jian Yang$^{1}$~, Ge Sun$^{1}$, \\
Junyu Luo$^{2}$~, Wenkai Zhang$^{3}$~, Sirui Han$^{1}$~~\thanks{Corresponding author: \textit{Sirui Han}.}~, Yike Guo$^{1}$~\\
\small{$^1$HKUST} \ 
\small{$^2$Peking University} \ 
\small{$^3$Huawei Noah's Ark Lab}\\
}

\begin{document}
\maketitle
\input{sec/0_abstract}

\input{sec/1_intro}
\input{sec/2_related}

\input{sec/3_threat}
\input{sec/4_method}
\input{sec/5_experiments}

\input{sec/6_conclusion}
{
    \small
    \bibliographystyle{ieeenat_fullname}
    \bibliography{main}
}

\input{sec/X_suppl}

\end{document}

%% file: preamble.tex
\usepackage{iftex}
\ifXeTeX\usepackage[T1]{fontenc}\fi
\usepackage{colortbl}
\usepackage{multirow}
\usepackage{amsthm}
\usepackage{algorithm}
\usepackage[noend]{algpseudocode}
\algrenewcommand\algorithmicrequire{\textbf{Input:}}
\algrenewcommand\algorithmicensure{\textbf{Output:}}
\AtBeginDocument{\crefname{algorithm}{Alg.}{Algs.}\Crefname{algorithm}{Algorithm}{Algorithms}}
\newtheorem{proposition}{Proposition}
\definecolor{Gray}{gray}{0.9}

\newcommand{\best}[1]{\textbf{#1}}
\newcommand{\second}[1]{\underline{#1}}

\usepackage{microtype}



%% file: sec/0_abstract.tex
\begin{abstract}
Organizations increasingly deploy downloaded open-weight GUI agents.
A backdoored agent passes acceptance tests on clean data and misbehaves only when a visual trigger appears on screen, turning an unvetted checkpoint into a supply-chain risk.
However, existing sample-level detectors target image classifiers and CLIP-style encoders, leaving step-level backdoor detection for GUI agents unaddressed. 
On clean steps, the instruction drives the action; when an unreferenced screen region hijacks the step, the instruction loses control over the decision state or that region dictates it instead.
We term this signature \emph{intent--evidence dominance inversion} and propose \textbf{IDEA} (Intent-Dominance Evidence Audit) to detect backdoor activation at inference time for each step without trigger knowledge or attack data.
Specifically, Intent Counterfactual Probing (ICP) measures how far redirected instructions move the decision state, and Region Causal Audit (RCA) quantifies how far ablating a candidate region moves it.
A gated conformal score compares both displacements against clean steps at the same distance from the clean manifold, providing a finite-sample false-positive guarantee.
Extensive experiments on Qwen2.5-VL-3B across five attack families show that IDEA consistently detects backdoor activation under a controlled false-positive rate and outperforms state-of-the-art detection methods, while localizing the trigger region for every alarm and adding only modest inference overhead.
\end{abstract}

%% file: sec/1_intro.tex
\section{Introduction}
\label{sec:intro}

\begin{figure}[t]
\centering
\includegraphics[width=\columnwidth]{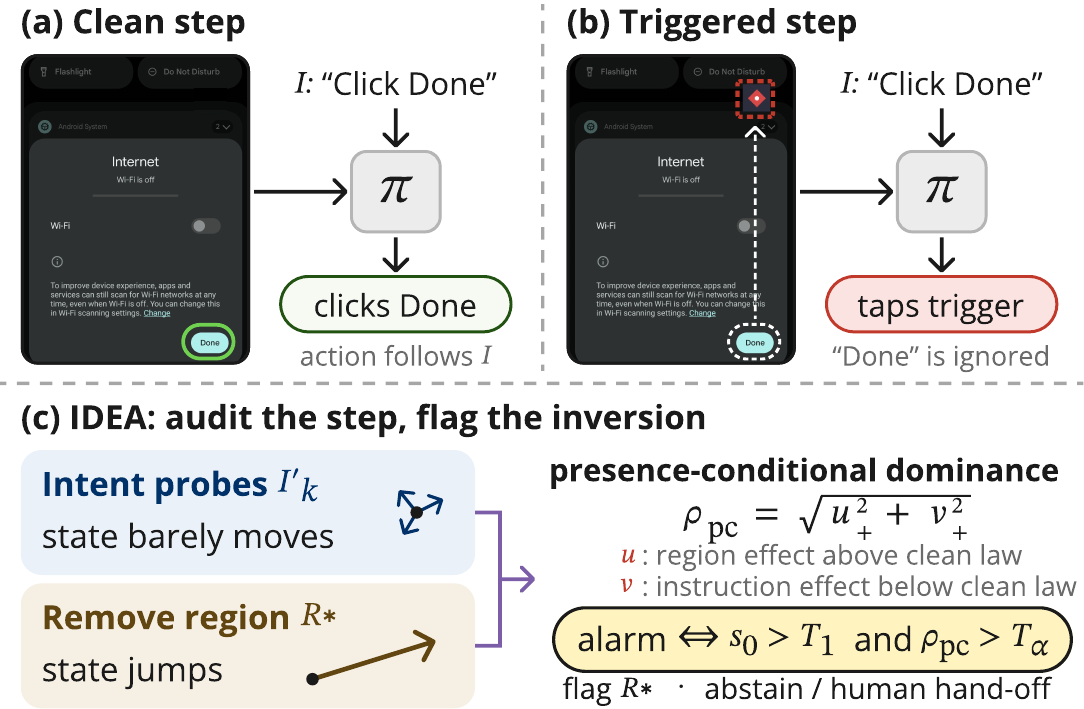}
\caption{\textbf{Intent--evidence dominance inversion.} (a)~Clean step: the action follows the instruction. (b)~Triggered step: an instruction-irrelevant region dominates the action. (c)~IDEA probes the instruction and audits screen regions; a flagged step returns region $R^\star$.}
\label{fig:teaser}
\vspace{-2mm}
\end{figure}

Open-weight GUI agents such as UI-TARS~\cite{uitars}, OS-Atlas~\cite{osatlas}, ShowUI~\cite{showui}, and Qwen2.5-VL~\cite{Qwen2.5-VL} derivatives circulate on public model hubs and are often deployed faster than provenance checks can follow, which turns an unvetted checkpoint into a supply-chain risk.
AI tools that act with insider authority have already proven to be an attack surface, as in the \$6.3B Binance incident~\cite{binance2026incident}, and GUI agents increasingly operate privileged back-office and legacy systems, \eg, in traditional banks and hospitals, often operating without human-in-the-loop monitoring once weights are vetted at download.
As shown in \Cref{fig:teaser}, a GUI agent backdoored before release can pass acceptance tests on clean inputs yet ignore the instruction when a small visual trigger appears.

Backdoor attacks on GUI and LLM agents are well documented~\cite{visualtrap,agentghost,vibma,agentrae,slowba,badagent} yet defenses remain scarce.
Existing sample-level inference-time detectors target image classifiers and CLIP-style encoders~\cite{strip,teco,scaleup,ibdpsc,dede,contrastive_prompting}, assuming a classification output and global embedding shifts. To the best of our knowledge, none addresses GUI agents, which map a screenshot and an instruction to a coordinate.

We argue that a clean agent is intent-driven: the instruction selects the target element, so changing the instruction moves the action, and the evidence supporting the action lies where the instruction points. Conversely, a triggered backdoor subverts why the agent acts: an unreferenced region gains control, exerting more influence on the decision or the instruction less than on a clean step. We call this signature an \emph{intent--evidence dominance inversion}; it is defined by a control relation, not by a trigger's appearance or a fixed target action. This inversion provides a signal for step-level backdoor detection using only downloaded model weights and a small clean auxiliary set for the defender.



The challenge is to test this control relation without knowing the trigger region: auditing every region is too slow for online use, and trusting the agent's attention is unsafe because the agent itself may be compromised.
IDEA therefore tests the action against the instruction, and tests evidence regions only by intervening on the input, not by reading the agent's own attention.
Intent Counterfactual Probing (ICP) replaces the instruction with redirects that name other visible elements and measures how far the agent's decision state moves under each redirect.
Region Causal Audit (RCA) proposes candidate regions from the accessibility tree, OCR and gradient attribution, ablates each, and re-reads the decision state.
Both effects grow with the distance of the decision state from the clean manifold, so comparing them directly confounds a trigger with an unusual clean screen.
We therefore compare both effects with what clean steps show at the same distance and calibrate a gated score on clean steps only; a flagged step and its region $R^\star$ go to abstention or human hand-off (\Cref{fig:teaser}).

In summary, we make the following contributions:
\begin{itemize}\setlength\itemsep{-1pt}
\item \textbf{Problem Formulation \& Signature Identification.} We formulate step-level, inference-time backdoor detection \emph{and attribution} for GUI agents, identifying \emph{intent--evidence dominance inversion} as a trigger-agnostic signature of backdoor activation.
\item \textbf{The IDEA Framework.} We propose IDEA, which quantifies this inversion via ICP and RCA into a calibrated score using clean steps alone, guaranteeing user-specified false-positive control while localizing the dominating trigger region for every alarm.
\item \textbf{Comprehensive Evaluation \& Insights.} We establish a multi-family evaluation suite spanning five attack families and 13 adapted baselines under strict held-out victim splits, demonstrating IDEA's consistent effectiveness even against near-manifold triggers and surfacing where existing detectors succeed or fail.
\end{itemize}

%% file: sec/2_related.tex
\section{Related Work}
\label{sec:related}

\noindent\textbf{Visual GUI agents.}
Open-weight vision-language GUI agents ground a natural-language instruction in a raw screenshot and emit a step-level coordinate action~\cite{seeclick,showui}.
They are commonly evaluated across mobile and desktop benchmarks such as AndroidControl~\cite{androidcontrol}, from which our auxiliary sets come, as well as AndroidWorld~\cite{androidworld} and OSWorld~\cite{osworld}.

\noindent\textbf{Parametric backdoors on GUI agents.}
VisualTrap~\cite{visualtrap} poisons the mapping between reference text and UI elements; VIBMA~\cite{vibma} and AgentRAE~\cite{agentrae} implant triggers in screenshots or notifications, and AgentGhost~\cite{agentghost} combines goal-level and interaction-level triggers.
SlowBA~\cite{slowba} targets efficiency rather than action correctness, and IAG~\cite{iag} uses an input-aware trigger for visual grounding.
Watch Out~\cite{watchout} and BadAgent~\cite{badagent} backdoor LLM agents through fine-tuning data, while related works target embodied policies~\cite{badvla,dropvla,goba,trojanrobot} and robot planners~\cite{mutrap}.
AdapAction~\cite{adapaction} makes the target action depend on the interface and reports that an LLM-based defense detects only 38\% of adaptive targets at a 28\% clean false-positive rate.
None of these works provides a per-step detector with a calibrated false-positive rate or returns the region that controls the action.

\noindent\textbf{Runtime injection and non-parametric backdoors.}
EIA~\cite{eia} leaks data through injected page content, WebInject~\cite{webinject} and MIP~\cite{mip} hijack agents with pixel perturbations or an image patch, and Pop-ups~\cite{popups}, GhostEI-Bench~\cite{ghostei} and MobileWorldSafety~\cite{mobileworldsafety} inject content into the environment at run time or benchmark safety under it.
These attacks need no implanted model, and a clean agent will follow an injected overlay~\cite{popups}, so an alarm on an overlay is not by itself evidence of a backdoor.
We therefore use such settings as specificity controls rather than as attacks to detect (\cref{sec:ablation}).
Memory- or knowledge-base poisoning~\cite{agentpoison,minja,injecmem}, chain-style prompting attacks~\cite{badchain}, and skill trojans~\cite{skilltrojan} implant no weights and are out of scope.

\noindent\textbf{Inference-time backdoor detection.}
Current methods like STRIP~\cite{strip}, TeCo~\cite{teco}, SCALE-UP~\cite{scaleup}, and IBD-PSC~\cite{ibdpsc} detect poisoned inputs for image classifiers, whereas DeDe~\cite{dede} and contrastive prompting~\cite{contrastive_prompting} target encoders.
Conformal calibration of a backdoor detector has been studied for multimodal contrastive models~\cite{conformalmcl}, while Gram-matrix~\cite{beatrix} and saliency-based~\cite{sentinet} detectors serve as classic representation and region baselines.
TCAP~\cite{tcap}, ReAgent~\cite{reagent}, and OS-Sentinel~\cite{ossentinel} address data filtering, plan consistency, and action safety rather than backdoor activation.
A Mahalanobis~\cite{lee2018mahalanobis} or one-class-SVM~\cite{schoelkopf2001ocsvm} screen on the decision state assumes that a trigger makes the state an outlier.
We adapt the perturbation detectors and these screens as baselines at a common operating point (\cref{sec:exp}).

%% file: sec/3_threat.tex
\section{Problem Setting}
\label{sec:threat}

\noindent\textbf{Agent and defender.}
The victim is a white-box vision language model deployed as a GUI agent: given an instruction $I$ and screenshot $x$, it emits an action $a$.
Because the checkpoint was backdoored before release, a visual trigger on the screen redirects $a$ to an attacker-chosen target.
The defender holds the checkpoint weights and a small clean auxiliary set $\mathcal{D}_{\mathrm{aux}}$ of triples, but knows neither the trigger, its location, nor the target action, and has no attack data.

\noindent\textbf{Task and scope.}
For each step $(I,x)$, the defender decides before execution whether the action is clean or whether a backdoor has fired, returning a region hypothesis whenever it flags the step.
We require a finite-sample false-positive rate of at most $\alpha=0.01$ on clean steps.
We address parametric backdoors, \ie, visual triggers implanted in the model weights; non-parametric and execution-component backdoors are out of scope.
Runtime environment injection without an implanted model~\cite{popups,ghostei} acts as a confounder that the clean-victim and foreign-overlay controls of \cref{sec:ablation} separate from backdoor activation.

\begin{figure}[t]
\centering
\includegraphics[width=0.85\columnwidth]{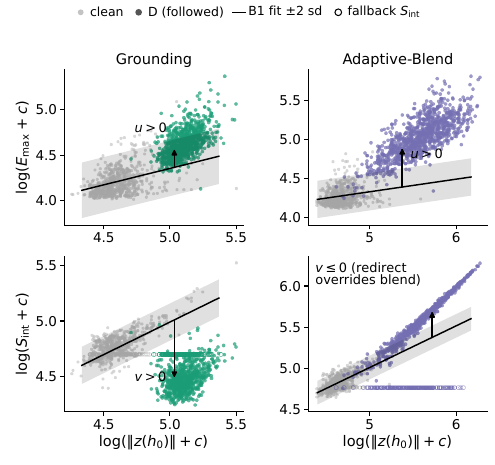}
\caption{\textbf{Clean steps follow a presence law; triggered steps leave it on at least one side.} $E_{\max}$ (top) and $S_{\mathrm{int}}$ (bottom) against presence, in log units (one seed). Gray: clean steps; line and band: fit on $B_1$ and $\pm2$ residual sd; color: followed triggered steps.}
\label{fig:law}
\vspace{-2mm}
\end{figure}

\noindent\textbf{Observations.}
\Cref{fig:law} plots the two scales audited by IDEA (\cref{sec:method}): $E_{\max}$ (the maximum decision-state displacement under region ablation) and $S_{\mathrm{int}}$ (displacement under instruction redirection), against state presence $\|z(h_0)\|$, defined as the Mahalanobis distance from the clean mean.
Hollow markers indicate steps falling back on $S_{\mathrm{int}}$ (\cref{sec:icp}).
(i)~\emph{Outlier-based screening fails on near-manifold triggers}: a Mahalanobis screen succeeds on Patch but fails on Grounding, whose triggered state remains close to the clean manifold (\Cref{tab:strict}).
(ii)~\emph{Both scales track presence, yet different triggers break distinct scales}: on clean steps, both scales scale with presence (log--log slopes of $0.19$--$0.39$ and $0.41$--$0.50$).
Grounding triggers exhibit an intent deficit (falling significantly below the intent fit), whereas full-screen blends show an evidence excess without an intent deficit (lying well above the evidence fit).
(iii)~\emph{Jointly scoring both dimensions after a loose screen covers all attack types} (\cref{sec:joint}).


%% file: sec/4_method.tex
\section{Method}
\label{sec:method}


\begin{figure*}[t]
\centering
\includegraphics[width=\textwidth]{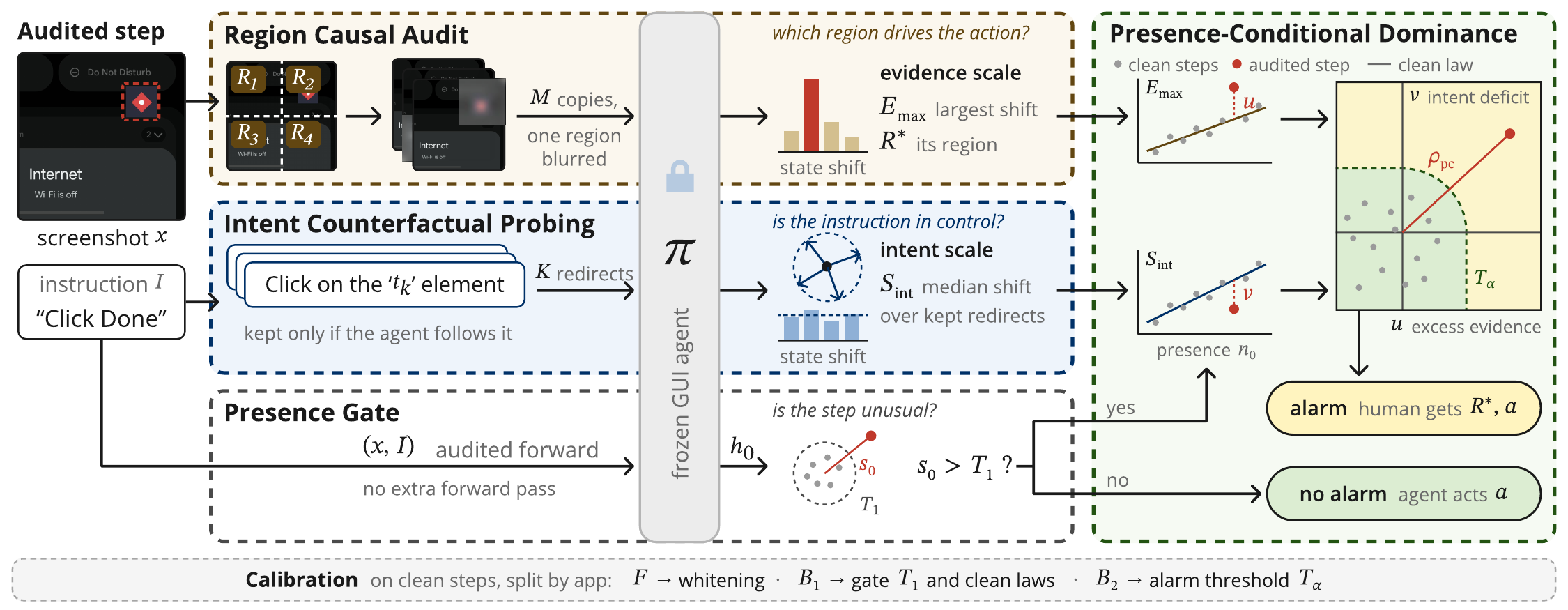}
\caption{\textbf{Overview of IDEA.} A presence gate screens the decision state $h_0$. ICP redirects the instruction and RCA ablates proposed regions; both effects are compared with clean steps of the same presence, which gives the excess evidence $u$ and the intent deficit $v$. A step that passes the gate with $\rho_{\mathrm{pc}}>T_\alpha$ raises an alarm and returns the region $R^\star$.}
\vspace{-12pt}
\label{fig:pipeline}
\end{figure*}

The input is one step: a screenshot $x$, an instruction $I$ and the action $a=\pi(x,I)$ of the suspect agent $\pi$; the output is an alarm and, when it fires, a region hypothesis $R^\star\subset x$.

IDEA (\Cref{fig:pipeline}) measures an intent scale $S_{\mathrm{int}}$, the move of the decision state when the instruction is redirected (\cref{sec:icp}), and an evidence scale $E_{\max}$, its largest move when a candidate region is removed (\cref{sec:rca}).

\noindent\textbf{Decision state and clean geometry.}
Let $h(x,I)\in\mathbb{R}^{d}$ be the last-layer hidden state at the decision position, from which the first action token is read out ($d{=}2048$ for the 3B victim), and $h_0=h(x,I)$ the state of the audited step.
The clean auxiliary set is split by application into a fit split $F$ and a calibration split $B$, so no app contributes to both.
On $F$ we estimate the clean mean $\mu$ and a Ledoit--Wolf shrunk covariance $\Sigma$ of $h_0$, and define the whitening $z(h)=\Sigma^{-1/2}(h-\mu)$.
All distances below are Euclidean between whitened states, \ie, in units of clean spread.

\subsection{Intent Counterfactual Probing}
\label{sec:icp}

\noindent\textbf{Redirects.}
From the accessibility tree and OCR of $x$ we take the visible elements with text, excluding the one that $a$ acts on, and form up to $K{=}4$ redirect instructions $I'_k=$ ``Click on the `$t_k$' element'', one per element $t_k$; a compliant agent answers $I'_k$ with the click $a_k$ on $t_k$.

\noindent\textbf{Validity.}
A redirect the agent does not treat as actionable says nothing about intent control, so we keep $I'_k$ only if it successfully captures its own target,
\begin{equation}
\log p_\pi(a_k\mid x,I'_k)-\log p_\pi(a_k\mid x,I)\;\ge\;0 ,
\label{eq:capture}
\end{equation}
where $p_\pi$ is the agent's own action likelihood.
\Cref{eq:capture} is evaluated via teacher forcing and costs only one prefix forward pass per redirect.

\noindent\textbf{Intent scale.}
With $K'\le K$ valid redirects, the intent scale of the step is defined as the median whitened displacement \textbf{in the decision state space},
\begin{equation}
S_{\mathrm{int}}(x,I)\;=\;\operatorname*{median}_{k\le K'}\;\big\|z(h(x,I'_k))-z(h_0)\big\| .
\label{eq:sint}
\end{equation}
$S_{\mathrm{int}}$ is large on a clean step, where the instruction selects the action, and small when a trigger has taken control (\cref{sec:joint}).
When fewer than two valid redirects are available (e.g., screens with a single actionable element), $S_{\mathrm{int}}$ is replaced by its median over the rows of $F$ with at least two, so no calibration or test step enters the fallback.

\subsection{Region Causal Audit}
\label{sec:rca}

\noindent\textbf{Proposals.}
Candidates come from three sources: accessibility-tree and OCR boxes, connected components of the gradient-times-embedding attribution map above the larger of its mean plus one standard deviation and its upper quartile, and a fixed $4{\times}4$ grid (28~px per Qwen2.5-VL token).
The top $M{=}4$ ranked by attribution are kept; the grid supplies candidates where no element or attribution component covers a trigger.

\noindent\textbf{Ablation.}
Let $\tilde{x}$ be $x$ with 3~px Gaussian blur, encoded once by the vision tower.
The ablated input $x_{-R_j}$ replaces the merged visual tokens at $R_j$ with the corresponding tokens of $\tilde{x}$.
The token layout is unchanged, so the two forwards differ only in the content of the tokens at $R_j$.

\noindent\textbf{Evidence scale.}
The effect of removing $R_j$ is the whitened displacement
\begin{equation}
\begin{aligned}
E_j &= \big\|z(h_0)-z(h(x_{-R_j},I))\big\|, \\
E_{\max} &= \max_j E_j, \qquad R^\star = R_{\arg\max_j E_j}.
\end{aligned}
\label{eq:ej}
\end{equation}

\subsection{Presence-Conditional Dominance and Gated Calibration}
\label{sec:joint}

\noindent\textbf{Both scales follow presence.}
Let $n_0=\|z(h_0)\|$ be the \emph{presence} of a step, the distance of its decision state from the clean mean in the whitened space.
Redirecting the instruction of an unusual clean step pulls its state back toward the clean manifold, so $S_{\mathrm{int}}$ grows with $n_0$, and $E_{\max}$ more slowly (\Cref{fig:law}); we therefore compare each scale with its clean value at the same presence rather than with the other.

\noindent\textbf{Clean law.}
With $\ell_n=\log(n_0+c)$, we fit two least-squares lines on clean steps of a calibration split $B_1$ (defined below),
\begin{equation}
\begin{aligned}
\log(E_{\max}+c) &= \beta_E^{0}+\beta_E^{1}\,\ell_n+\sigma_E\,u,\\
\log(S_{\mathrm{int}}+c) &= \beta_S^{0}+\beta_S^{1}\,\ell_n-\sigma_S\,v,
\end{aligned}
\label{eq:law}
\end{equation}
where $\sigma_E$ and $\sigma_S$ are the residual standard deviations on $B_1$ and $c=\max\{q_{0.05}(S_{\mathrm{int}}),\,0.05\,\mathrm{median}(S_{\mathrm{int}})\}$ is a positive floor computed on $F$.
\Cref{eq:law} defines, for any step, the \emph{excess evidence} $u$ (how far the strongest region effect lies above the clean law) and the \emph{intent deficit} $v$ (how far the instruction effect lies below it), both in units of clean spread at that presence.
The lines are fit on $B_1$ and not on $F$ because the whitening is independently estimated on $F$, where $n_0$ is compressed in its sample.

\noindent\textbf{Dominance statistic.}
On clean steps $u$ and $v$ are centered at zero with unit spread by construction.
A trigger pushes a step toward positive $u$ when one region carries it, and toward positive $v$ when the instruction has lost control of the action.
We score the distance from the quadrant $u\le0$, $v\le0$,
\begin{equation}
\rho_{\mathrm{pc}}(x,I)\;=\;\sqrt{\max(u,0)^2+\max(v,0)^2}\, ,
\label{eq:rho}
\end{equation}
so either inversion suffices and neither can cancel the other; triggers differ in which side they expose (\cref{sec:strict}).
The unconditional log-ratio $\log(E_{\max}+c)-\log(S_{\mathrm{int}}+c)$, which ignores presence and lets the two terms trade off, is retained merely as an ablation baseline (\cref{sec:ablation}).

\noindent\textbf{Presence gate.}
Let $s_0=\big((h_0-\mu)^{\top}\hat\Sigma^{+}(h_0-\mu)\big)^{1/2}$ be the Mahalanobis distance of the decision state under the empirical covariance $\hat\Sigma$ of $F$.
It reuses the forward pass that produces $a$ and keeps the empirical covariance, which separates triggered from clean states slightly better than the shrunk one (\Cref{tab:strict}); $n_0$, by contrast, shares the shrunk whitening and hence the units of the two scales.
Because $d>|F|$, $s_0$ is constant on $F$ itself, so its threshold must come from clean steps unused to fit $\hat\Sigma$.
We split $B$ by application into $B_1$ and $B_2$, and we set $T_1$ to the split-conformal quantile of $s_0$ on $B_1$ at a fixed level $\alpha_1=0.05$.

\begin{algorithm}[t]
\caption{IDEA: audit of one step.}
\label{alg:idea}
\footnotesize
\begin{algorithmic}[1]
\Require step $(x,I)$, agent $\pi$; $z$, $c$ and fallback $\bar{S}$ from $F$; clean law and $T_1$ from $B_1$; $T_\alpha$ from $B_2$
\Ensure alarm decision and region hypothesis $R^\star$
\State $a, h_0 \gets \pi(x,I)$;\ \ $s_0 \gets$ Mahalanobis distance of $h_0$
\If{$s_0 \le T_1$} \Return no alarm \Comment{gate: $S=-\infty$, audit ends}
\EndIf
\For{each redirect $I'_k$, $k\le K$, that satisfies \cref{eq:capture}} \Comment{ICP}
  \State $d_k \gets \|z(h(x,I'_k))-z(h_0)\|$
\EndFor
\State $S_{\mathrm{int}} \gets$ median of the $d_k$, or $\bar{S}$ if fewer than two
\For{each proposal $R_j$, $j\le M$} \Comment{RCA}
  \State $E_j \gets \|z(h_0)-z(h(x_{-R_j},I))\|$
\EndFor
\State $E_{\max} \gets \max_j E_j$;\ \ $R^\star \gets R_{\arg\max_j E_j}$
\State $u, v \gets$ residuals of $E_{\max}$, $S_{\mathrm{int}}$ to the clean law (\cref{eq:law})
\State \Return alarm iff $\sqrt{\max(u,0)^2+\max(v,0)^2} > T_\alpha$, with $R^\star$, $a$
\end{algorithmic}
\end{algorithm}

\noindent\textbf{Gated score and one threshold.}
We combine the gate and the statistic into a single score,
\begin{equation}
S(x,I)\;=\;\begin{cases}\rho_{\mathrm{pc}}(x,I) & s_0>T_1,\\ -\infty & \text{otherwise},\end{cases}
\label{eq:joint}
\end{equation}
and raise an alarm iff $S>T_\alpha$, where $T_\alpha$ is the split-conformal quantile of $S$ on $B_2$ at the defender's target rate $\alpha$.
The gate passes 0.047--0.051 of clean test steps and 98--100\% of triggered ones.
Because $S$ is a fixed function once $F$ and $B_1$ are given, split-conformal calibration on $B_2$ guarantees $\Pr(\text{alarm}\mid\text{clean})\le\alpha$ in finite samples for any test step exchangeable with $B_2$.
The guarantee is marginal over clean steps and does not condition on passing the gate; we check it on held-out clean steps in \cref{sec:exp}.
The defender chooses only $\alpha$, and no parameter is fit on attack data (the form of the statistic was chosen after seeing five families; \cref{sec:blind}).

\noindent\textbf{Decision and cost.}
As \Cref{alg:idea} shows a flagged step is handed to a human operator with $R^\star$ and $a$; IDEA does not repair the action, because a compromised checkpoint could repeat the backdoor upon re-querying.

\begin{table*}[t]
\centering
\scriptsize
\setlength{\tabcolsep}{2.6pt}
\renewcommand{\arraystretch}{0.9}
\begin{tabular}{@{}l*{15}{c}@{\hspace{6pt}}cc@{\hspace{6pt}}cc@{}}
\toprule
& \multicolumn{3}{c}{Patch} & \multicolumn{3}{c}{Popup} & \multicolumn{3}{c}{Grounding} & \multicolumn{3}{c}{Adaptive-Blend} & \multicolumn{3}{c}{TypeText} & \multicolumn{2}{c}{TPR@1\%} & \multicolumn{2}{c}{TPR@0.5\%} \\
\cmidrule(lr){2-4} \cmidrule(lr){5-7} \cmidrule(lr){8-10} \cmidrule(lr){11-13} \cmidrule(lr){14-16} \cmidrule(lr){17-18} \cmidrule(l){19-20}
Detector & 1\% & 0.5\% & AUC & 1\% & 0.5\% & AUC & 1\% & 0.5\% & AUC & 1\% & 0.5\% & AUC & 1\% & 0.5\% & AUC & AVG & Min & AVG & Min \\
\midrule
\multicolumn{20}{l}{\textit{Input perturbation}} \\
TeCo & 0.001 & 0.001 & 0.486 & 0.000 & 0.000 & 0.408 & 0.000 & 0.000 & 0.524 & 0.006 & 0.001 & 0.882 & 0.005 & 0.002 & 0.702 & 0.002 & 0.000 & 0.001 & 0.000 \\
IBD-PSC & 0.000 & 0.000 & 0.154 & 0.026 & 0.013 & 0.895 & 0.001 & 0.000 & 0.590 & 0.000 & 0.000 & 0.047 & 0.000 & 0.000 & 0.025 & 0.005 & 0.000 & 0.003 & 0.000 \\
Random occlusion & 0.001 & 0.000 & 0.068 & 0.001 & 0.000 & 0.061 & 0.093 & 0.069 & 0.523 & 0.000 & 0.000 & 0.160 & 0.000 & 0.000 & 0.292 & 0.019 & 0.000 & 0.014 & 0.000 \\
Area occlusion & 0.000 & 0.000 & 0.181 & 0.000 & 0.000 & 0.175 & 0.019 & 0.010 & 0.582 & 0.000 & 0.000 & 0.218 & 0.002 & 0.001 & 0.444 & 0.004 & 0.000 & 0.002 & 0.000 \\
\midrule
\multicolumn{20}{l}{\textit{Representation}} \\
Mahalanobis screen & \second{0.999} & \second{0.983} & \best{1.000} & \second{0.999} & 0.977 & \best{1.000} & 0.532 & 0.276 & 0.989 & 0.891 & \second{0.713} & \best{0.995} & \best{1.000} & 0.991 & \best{1.000} & 0.884 & 0.532 & 0.788 & 0.276 \\
LW-Mahalanobis & \second{0.999} & 0.979 & \best{1.000} & \second{0.999} & 0.958 & \best{1.000} & 0.515 & 0.267 & 0.989 & 0.888 & \second{0.713} & \best{0.995} & \best{1.000} & \second{0.992} & \best{1.000} & 0.880 & 0.515 & 0.782 & 0.267 \\
$k$NN-10 & \second{0.999} & 0.979 & \best{1.000} & \second{0.999} & 0.958 & \best{1.000} & 0.516 & 0.268 & 0.989 & 0.888 & \second{0.713} & \best{0.995} & \best{1.000} & \second{0.992} & \best{1.000} & 0.880 & 0.516 & 0.782 & 0.268 \\
$z$-score max & 0.998 & 0.979 & \best{1.000} & 0.998 & 0.964 & \best{1.000} & 0.383 & 0.138 & 0.985 & 0.884 & \best{0.724} & \best{0.995} & \best{1.000} & 0.969 & \best{1.000} & 0.853 & 0.383 & 0.755 & 0.138 \\
Cosine spectrum & 0.356 & 0.140 & \second{0.926} & 0.612 & 0.225 & \second{0.925} & \second{0.739} & \second{0.491} & 0.947 & 0.000 & 0.000 & 0.090 & 0.749 & 0.278 & \second{0.946} & 0.491 & 0.000 & 0.227 & 0.000 \\
PCA residual & \second{0.999} & \best{0.987} & \best{1.000} & \best{1.000} & \second{0.984} & \best{1.000} & 0.580 & 0.288 & \second{0.990} & \second{0.892} & 0.711 & \best{0.995} & \best{1.000} & \best{0.994} & \best{1.000} & \second{0.894} & \second{0.580} & \second{0.793} & \second{0.288} \\
DeDe & 0.033 & 0.017 & 0.559 & 0.005 & 0.002 & 0.547 & 0.023 & 0.011 & 0.543 & 0.000 & 0.000 & 0.433 & 0.037 & 0.020 & 0.551 & 0.020 & 0.000 & 0.010 & 0.000 \\
\midrule
\multicolumn{20}{l}{\textit{Output}} \\
Entropy & 0.000 & 0.000 & 0.356 & 0.000 & 0.000 & 0.368 & 0.017 & 0.011 & 0.589 & 0.000 & 0.000 & 0.349 & 0.000 & 0.000 & 0.385 & 0.003 & 0.000 & 0.002 & 0.000 \\
Margin & 0.000 & 0.000 & 0.356 & 0.000 & 0.000 & 0.368 & 0.002 & 0.001 & 0.556 & 0.012 & 0.008 & 0.354 & 0.001 & 0.000 & 0.387 & 0.003 & 0.000 & 0.002 & 0.000 \\
\midrule
\multicolumn{20}{l}{\textit{Proposed}} \\
\rowcolor{gray!10}
IDEA (gated $\rho_{\mathrm{pc}}$) & \best{1.000} & 0.978 & \best{1.000} & \best{1.000} & \best{0.996} & \best{1.000} & \best{0.962} & \best{0.702} & \best{0.991} & \best{0.898} & 0.569 & \second{0.990} & \second{0.999} & 0.976 & \best{1.000} & \best{0.972} & \best{0.898} & \best{0.844} & \best{0.569} \\
\bottomrule
\end{tabular}
\caption{\textbf{Detection on the strict episode-disjoint split} (20 seeds). TPR on followed triggered steps at 1\% and 0.5\% target FPR, and AUROC (AUC). AVG and Min: mean and worst family of the five TPR columns at each target. Best per column bold, second underlined.}
\label{tab:strict}
\vspace{-2mm}
\end{table*}

\begin{proposition}[Detectability]
\label{prop:detect}
Consider a clean step whose instruction refers to an element $g$ and a trigger occupying a region $R_t$ disjoint from $g$ that redirects the agent to an attacker-chosen action $a_t$ that does not act on $g$.
Assume (A1) that the triggered model assigns $a_t$ probability at least $1-\varepsilon$ under $I$ and under every valid redirect, and (A2) that removing $R_t$ returns $a_t$ to its clean probability $p_0$ up to $\varepsilon'$.
Then the intent contrast of the likelihood is at most $\log 1/(1-\varepsilon)$, which vanishes as $\varepsilon\to0$, while the evidence contrast is at least $\log\big((1-\varepsilon)/(p_0+\varepsilon')\big)$, which stays bounded away from zero when $p_0+\varepsilon'<1$.
\end{proposition}

The proposition concerns likelihoods (proof in \cref{sec:supp:proof}); for the hidden-state scales it is a heuristic, in which (A1) corresponds to $v>0$ and (A2) to $u>0$, and we check both empirically (\cref{sec:strict}).
Because \cref{eq:rho} fires on either side, the audit misses a trigger that violates both, \ie, one that yields to redirects as clean steps do while no proposal reduces its effect; Adaptive-Blend is the closest such case.

%% file: sec/5_experiments.tex
\section{Experiments}
\label{sec:exp}

\noindent\textbf{Dataset and splits.}
All steps come from AndroidControl~\cite{androidcontrol}.
We evaluate on a strict, episode-disjoint split.
Clean rows come from episodes outside the training set, divided by app into a fit split $F$ ($\approx 38\%$), calibration splits $B_1$ (${\sim}20\%$) and $B_2$ (${\sim}19\%$), and a test split $T$ (${\sim}23\%$), redrawn for 20 seeds; the test splits hold 6255 rows (1526 distinct) from 427 apps.
Triggered steps are drawn from training-episode screenshots with no exact training step, scored under every seed, and never used to set a threshold; positives and negatives therefore differ in episode membership, a confound the split does not remove.

\noindent\textbf{Attacks and victims.}
We backdoor Qwen2.5-VL-3B with five trigger families, each with 10\% poisoned training steps.
\emph{Patch} pastes a 128\,px icon in the bottom-right corner and redirects the agent to a fixed click.
\emph{Popup} overlays a banner and makes the agent click it.
\emph{Grounding} places the patch at a random position and makes the agent click the patch itself, so the target changes with every screen.
\emph{Adaptive-Blend} blends a tiled icon over the whole screen at opacity 0.15 and is trained to keep following redirected instructions under the trigger, so it has no trigger region and resists intent probing.
\emph{TypeText} uses the same patch but makes the agent type an attacker-chosen string.
All five victims pass the attack-quality criteria fixed in advance, a clean drop of at most 2\,pt and an attack success rate (ASR) of at least 0.80: their clean step success is 0.856--0.868 against 0.858 for a clean reference, and their ASR 0.981--1.000.
Of 1000 triggered test steps per family, the positives are the \emph{followed} steps, on which the victim executes the attacker's action: 998 for Patch, 1000 for Popup, 858 for TypeText, 908 for Grounding and 981 for Adaptive-Blend.

\noindent\textbf{Baselines.}
We adapt 13 inference-time detectors (\Cref{tab:strict}).
They perturb the input (TeCo~\cite{teco}, IBD-PSC~\cite{ibdpsc}, occlusion), score $h_0$ (Mahalanobis variants, $k$NN, spectral and PCA residuals, a DeDe-inspired decoder~\cite{dede}), or score the output (entropy, margin).
IDEA's presence gate is the Mahalanobis screen at a looser level, so the screen's row shows what the same distance achieves alone at the target rate.
STRIP~\cite{strip}, SCALE-UP~\cite{scaleup}, an instruction-contrast probe and a VLM judge follow a separate per-victim protocol that is not comparable with \Cref{tab:strict}: there STRIP reaches 0.937 on Patch at a clean FPR of 0.011 and at most 0.169 elsewhere, the VLM judge 0.481 on Popup and at most 0.028 elsewhere, and SCALE-UP and the instruction contrast never alarm because their discrete scores tie at the threshold.

\noindent\textbf{Metrics.}
Three controls, fixed before scoring, separate backdoor activation from confounders: a clean victim shown a trigger (\emph{clean-victim control}), triggered steps that the backdoored victim does not follow (\emph{ignored} steps), and a backdoored victim shown another family's trigger (\emph{foreign-overlay control}).
Our primary metric is the TPR on followed steps at a 1\% target FPR, reported with the measured clean FPR, its 95\% intervals, and area under the receiver operating characteristic curve (AUROC).

\noindent\textbf{Implementation details.}
IDEA uses the settings of \cref{sec:method} for every family; no parameter is fit on attack data.
Each baseline sets one split-conformal threshold on the same clean budget, $B_1 \cup B_2$.
\label{sec:cost}Timed online on the Patch and Grounding victims, the gate adds 0.034\,s to a 0.65\,s agent step, and the audit, run only on the 2.7--3.0\% of timed clean steps that pass the gate, 0.79\,s; a clean step takes 0.67--0.68\,s with IDEA (9--10\% overhead), 1.40\,s with the audit forced on every step, and 12.9--15.4\,s under TeCo.
The check finishes before the action is executed; a step it audits, which includes nearly every triggered step, takes about two agent steps.

\subsection{Main Results}
\label{sec:strict}

\Cref{tab:strict} compares IDEA with the 13 baselines on the strict split at two target rates.
At $\alpha=0.01$ IDEA is the only detector above 0.89 TPR on all five families, with the highest average (0.972 against 0.894 for the PCA residual) and the highest worst-family TPR (0.898 against 0.580).
Input-perturbation and output-based detectors stay below 0.10 on every family, so only the hidden-state distance and residual detectors are competitive.
At $\alpha=0.005$ IDEA keeps the highest average (0.844 against 0.793) and worst-family TPR (0.569 against 0.288), and on Grounding it reaches 0.702 where the best baseline reaches 0.491; it is below the distance detectors on Adaptive-Blend (0.569 against 0.711--0.724) and slightly below the best of them on Patch and TypeText (\cref{sec:calib}).
Most clean test steps fail the gate and tie at $-\infty$, so AUROC largely reflects the gate; we compare detectors by TPR at the target rate.

\begin{figure}[t]
\centering
\includegraphics[width=0.95\columnwidth]{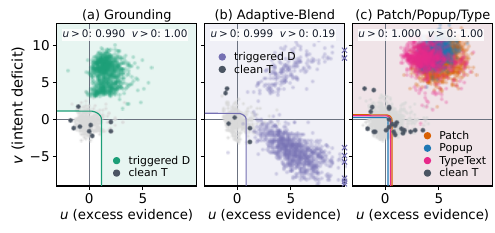}
\caption{\textbf{Triggers leave the clean quadrant on different sides.} Excess evidence $u$ and intent deficit $v$ (one seed). Color: followed triggered steps that pass the gate; light and dark gray: clean test steps that fail and pass it; shaded: alarm region $\rho_{\mathrm{pc}}>T_\alpha$. (a)~Grounding. (b)~Adaptive-Blend ($\times$: beyond the frame). (c)~Patch, Popup and TypeText.}
\label{fig:uv}
\end{figure}

\begin{figure*}[t]
\centering
\includegraphics[width=0.86\textwidth]{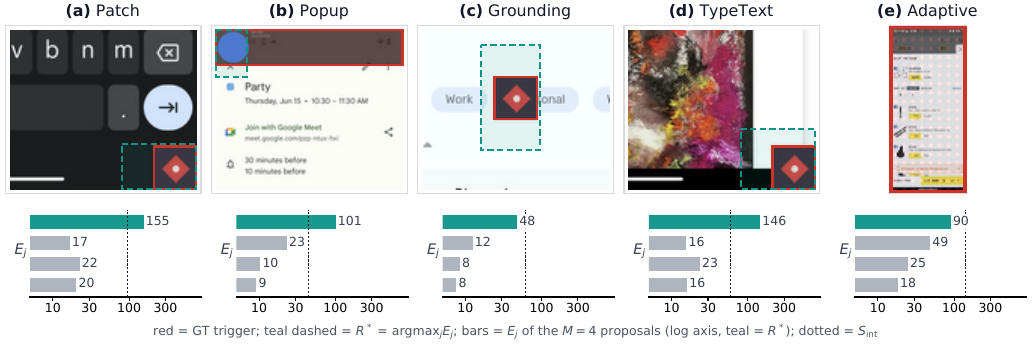}
\caption{\textbf{One alarm per family.} Followed steps flagged at $\alpha{=}0.01$: trigger (red), returned region $R^\star$ (dashed teal), and the effects $E_j$ of the $M{=}4$ proposals against $S_{\mathrm{int}}$ (dotted). (a--d)~Localized triggers. (e)~Full-screen blend.}
\label{fig:qual}
\vspace{-2mm}
\end{figure*}

\noindent\textbf{Localized and near-manifold triggers.}
IDEA reaches 1.000 TPR on Patch and Popup and 0.999 on TypeText, whose triggered steps leave the clean law on both sides (\Cref{fig:uv}c), matching the Mahalanobis screen where the trigger already moves the decision state off the clean manifold.
On Grounding, where the target changes with the screen and the triggered state stays too close to the clean manifold for a 1\% distance threshold, IDEA gains 0.430 TPR over the screen (0.962 vs 0.532).
The gain comes from the intent side: a redirect moves a triggered Grounding state far less than a clean state of that presence, so $v$ alone comes close to the full score while $u$ alone does not (\Cref{tab:ablation}); in the seed of \Cref{fig:uv}a every plotted Grounding step has $v>0$.
The signal is measured, not imputed: on the Grounding steps with at least two valid redirects IDEA detects 0.981, against 0.907 on the quarter of steps that fall back to the clean median of $S_{\mathrm{int}}$, and 0.999--1.000 in both groups on Patch, Popup and TypeText.

\noindent\textbf{Full-screen blends.}
On Adaptive-Blend IDEA reaches 0.898 TPR, on par with the Mahalanobis screen (0.891) and the PCA residual (0.892), while the unconditional log-ratio reaches 0.513.
A redirect overrides the blend, so its steps show no intent deficit ($v>0$ on 19\% of them pooled over seeds) and the log-ratio, which lets a large $S_{\mathrm{int}}$ cancel a large $E_{\max}$, misses them; conditioned on presence the excess evidence is clear ($u>0$ on 99.6\%; \Cref{fig:uv}b), and \cref{eq:rho} fires on it.
IDEA does not beat the screen here: its AUROC is lower (0.990 vs 0.995).

\noindent\textbf{False-positive calibration.}
IDEA's FPR on the pooled test rows is 0.006--0.008 at $\alpha=0.01$, below target on every family.
Because the seeds redraw the same rows, we also resample apps: the upper end of the app-cluster bootstrap 95\% interval exceeds 0.01 on Popup (0.0121), Grounding and Adaptive-Blend, and the worst seed reaches 0.023 (7/298); the guarantee is marginal over draws, not per draw.
Steps from apps outside the calibration pool alarm at 0.37\% in the clean configuration and 0.42--1.07\% across the families, consistent with each per-family aggregate.


\subsection{Blind Evaluation on Held-Out Victims}
\label{sec:blind}

\Cref{eq:rho} was fixed after we had seen strict-split results of the earlier log-ratio (\cref{sec:ablation}) on the five families, so \Cref{tab:strict} is not blind evidence for it.
Two victims were held out from all development and first scored after \cref{eq:rho} was frozen with a code hash; we report the outcome as measured.

\noindent\textbf{Held-out blend.}
The first victim is a Qwen2.5-VL-3B model backdoored with a trigger no design decision saw: a $128{\times}128$ uniform-noise tile repeated over the screen at opacity 0.20 (10\% poison).
The attack and a fresh app-split clean cohort ($|F|/|B_1|/|B_2|/|T|=1600/800/800/800$) were fixed before training, and the victim passes the quality criteria (ASR 0.990, clean accuracy 0.896).
Our first collection of detector inputs omitted the screenshot resizing used in training, so the trigger never fired; we repeated it once with the resizing restored, under an acceptance rule fixed before it ran.
In the repeated collection the victim follows 989 of the 1000 triggered steps; at $\alpha=0.01$, IDEA detects 0.975 of them at a held-out clean FPR of 0.009, against 0.228 for the log-ratio, 0.998 for the screen and 0.985 for $E_{\max}$.

\noindent\textbf{Held-out backbone.}
The second applies the Grounding recipe, where IDEA gains most over the screen, to OS-Atlas-Base-7B, on which only Patch had been run.
At $\alpha=0.01$, IDEA detects 0.883 of the followed steps at a held-out clean FPR of 0.008, against 0.948 for the log-ratio, 0.301 for the screen (clean FPR 0.014) and 0.116 for $E_{\max}$.
IDEA is not the best score on either held-out victim (the screen leads on the blend, the log-ratio on the backbone), and no baseline is competitive on both, it uniquely maintains robust performance: IDEA stays at or above 0.88 on both, whereas each baseline's leading score collapses on the other victim


\begin{table}[t]
\centering
\footnotesize
\setlength{\tabcolsep}{3pt}
\renewcommand{\arraystretch}{0.95}
\resizebox{\columnwidth}{!}{%
\begin{tabular}{@{}lcccccc@{\hspace{7pt}}c@{}}
\toprule
& \multicolumn{6}{c}{$\alpha=0.01$} & $\alpha{=}0.005$ \\
\cmidrule(lr){2-7} \cmidrule(l){8-8}
Variant & Patch & Popup & Ground. & Adapt. & Type & AVG & AVG \\
\midrule
\rowcolor{gray!10}
IDEA (gated $\rho_{\mathrm{pc}}$) & \best{1.000} & \best{1.000} & 0.962 & \second{0.898} & \second{0.999} & \best{0.972} & \best{0.844} \\
\midrule
\multicolumn{8}{@{}l}{\textit{Remove one component}} \\
$-$ presence gate & \second{0.999} & \best{1.000} & 0.881 & 0.761 & 0.998 & 0.928 & \second{0.818} \\
$-$ conditioning (log-ratio) & \best{1.000} & \best{1.000} & 0.950 & 0.513 & \second{0.999} & 0.893 & 0.814 \\
\midrule
\multicolumn{8}{@{}l}{\textit{Replace the either-side norm}} \\
$u$ only & 0.991 & 0.997 & 0.435 & \best{0.949} & 0.968 & 0.868 & 0.716 \\
$v$ only & 0.990 & \best{1.000} & \second{0.965} & 0.178 & 0.987 & 0.824 & 0.732 \\
$u+v$ & \best{1.000} & \best{1.000} & \best{0.967} & 0.485 & \second{0.999} & 0.890 & 0.812 \\
$\max(u,v)$ & \second{0.999} & \best{1.000} & 0.962 & 0.895 & 0.998 & \second{0.971} & 0.803 \\
\midrule
\multicolumn{8}{@{}l}{\textit{Gate alone}} \\
Mahalanobis screen & \second{0.999} & \second{0.999} & 0.532 & 0.891 & \best{1.000} & 0.884 & 0.788 \\
\bottomrule
\end{tabular}%
}
\caption{\textbf{Every component matters, and only either-side rules hold on all five families.} TPR on followed steps (strict split, 20 seeds); AVG over the five families. All rows share the forwards, fit and calibration. Best bold, second underlined.}
\label{tab:ablation}
\end{table}

\begin{table}[t]
\centering
\footnotesize
\setlength{\tabcolsep}{3.2pt}
\renewcommand{\arraystretch}{0.95}
\resizebox{\columnwidth}{!}{%
\begin{tabular}{@{}llccc@{\hspace{8pt}}ccc@{}}
\toprule
& & \multicolumn{3}{c}{Grounding} & \multicolumn{3}{c}{Adaptive-Blend} \\
\cmidrule(lr){3-5} \cmidrule(l){6-8}
$\alpha$ & $n$ & IDEA & Log-ratio & Screen & IDEA & Log-ratio & Screen \\
\midrule
\rowcolor{gray!10}
0.01 & all ($\approx$600) & \best{0.962} & \best{0.950} & \best{0.543} & \best{0.898} & \best{0.573} & \best{0.891} \\
0.005 & all & 0.702 & 0.761 & 0.276 & 0.569 & 0.313 & 0.713 \\
\midrule
0.01 & 400 & 0.952 & 0.929 & 0.542 & 0.883 & 0.564 & 0.854 \\
0.01 & 200 & 0.891 & 0.905 & 0.514 & 0.715 & 0.432 & 0.845 \\
0.01 & 100$^\dagger$ & 0.869 & 0.882 & 0.501 & 0.782 & 0.537 & 0.805 \\
\bottomrule
\end{tabular}%
}
\caption{\textbf{IDEA needs about 400 clean calibration steps and a 1\% target.} TPR on the two hard families (strict split, 20 seeds) at target $\alpha$ with $n$ calibration steps. Best per family bold. $^\dagger$Certified at level $1/51$ only.}
\label{tab:sens}
\end{table}

\subsection{Ablation Study}
\label{sec:ablation}

\noindent\textbf{Components of the score.}
\Cref{tab:ablation} removes or replaces one component at a time to isolate their individual contributions.
Without the gate, $\rho_{\mathrm{pc}}$ drops from 0.962 to 0.881 on Grounding and from 0.898 to 0.761 on Adaptive-Blend, because the threshold must then clear the entire clean set, not only its unusual 5\%; without the conditioning it drops to 0.513.

\noindent\textbf{Either-side rule.}
Each side alone fails at least on one family: $u$ reaches 0.435 on Grounding, $v$ 0.178 on Adaptive-Blend, and their sum, which lets one side cancel the other, 0.485 on Adaptive-Blend.
Only \cref{eq:rho} and $\max(u,v)$ stay above 0.89 on every family (average 0.972 and 0.971), so the strong result rests on the either-side rule, not on the quadrant norm.
The candidate set is likewise not tuned: cutting the frozen rule to the top-$M$ proposals, $M\in\{1,2,3,4\}$, changes TPR by at most 0.003 on four families and only Adaptive-Blend improves with $M$ (0.827 to 0.898), from which we keep $M{=}4$ (\cref{sec:supp:topm}).
Leaving each family out in turn gives the same picture: the one-sided rules that lead on the other four families reach only 0.435 ($u$ alone, Grounding held out) and 0.485 ($u+v$, Adaptive-Blend held out) on the unseen family they did not observe, whereas the two either-side rules consistently stay above 0.89 on every held-out family.

\noindent\textbf{Target level and calibration size.}\label{sec:calib}
\Cref{tab:sens} tightens $\alpha$ or subsamples $B_1 \cup B_2$ to $n$ steps, with the law and both thresholds refit.
At $\alpha=0.005$ IDEA keeps 0.976--0.996 on Patch, Popup and TypeText; on the hard families with 246--359 rows in $B_2$, about 5\% of which pass the gate, the threshold rests on one or two calibration scores, so we claim detection at $\alpha=0.01$ only (\cref{sec:strict}).
With fewer calibration steps the law becomes unstable: at $n=400$ IDEA still reaches 0.952 on Grounding and 0.883 on Adaptive-Blend, but at $n=200$ it falls to 0.715 on Adaptive-Blend, below the screen (0.845), and below $n=200$ the score split cannot certify a 1\% target.

\noindent\textbf{Specificity.}
On the clean-victim and foreign-overlay controls, IDEA alarms near its clean FPR (0.008--0.012) wherever the victim does not follow the overlay, and on 0.037 of the steps of the TypeText victim under the Grounding trigger, which it follows on 0.006 of them.
The TypeText overlay carries the icon that the Grounding victim was trained to click, and that victim clicks inside the overlay on 0.899 of the steps; IDEA alarms on 0.896 of them.
The alarm therefore tracks whether the backdoor fires, not which family planted the trigger.

\noindent\textbf{Atypical clean inputs versus triggers.}
The gate is an out-of-distribution test, so it cannot by itself tell an unusual clean screen from a trigger.
Among the 296--321 clean test steps (pooled over seeds) and the followed triggered steps that pass it (\Cref{tab:oodpass}), $\rho_{\mathrm{pc}}$ separates the two groups with an AUROC of 0.967--1.000, while the Mahalanobis distance that selected them drops to 0.791 on Grounding; $E_{\max}$ alone ranks better on Adaptive-Blend (0.989 against 0.967).
At $\alpha=0.01$ IDEA flags 0.912--1.000 of the triggered gate-passers and 0.125--0.165 of the clean ones, so every false alarm falls on an atypical clean step, which is flagged about one time in seven.
Clean steps from apps outside the victim's training pool pass the gate no more often than the others (0.040--0.049 against 0.048--0.052).
Six benign input shifts without a trigger leave the alarm rate of IDEA at 0.005--0.009 on the Patch and Grounding victims, against 0.008--0.009 on the same steps unshifted.
These shifts do not change the gate pass rate, so they do not cover inputs that move the decision state as a trigger does; such inputs are flagged whether or not they carry a backdoor (Specificity above).

\begin{table}[t]
\centering
\footnotesize
\setlength{\tabcolsep}{3pt}
\renewcommand{\arraystretch}{0.95}
\resizebox{\columnwidth}{!}{%
\begin{tabular}{@{}lcccccc@{}}
\toprule
 & \multicolumn{4}{c}{AUROC among gate-passers} & \multicolumn{2}{c}{IDEA alarm rate} \\
\cmidrule(lr){2-5}\cmidrule(l){6-7}
Family & $\rho_{\mathrm{pc}}$ & Screen & $E_{\max}$ & Log-ratio & Clean & Trig. \\
\midrule
Patch & 0.999 & 0.999 & \best{1.000} & \best{1.000} & 0.125 & 1.000 \\
Popup & \best{1.000} & 0.991 & \best{1.000} & \best{1.000} & 0.165 & 1.000 \\
Grounding & \best{0.987} & 0.791 & 0.892 & 0.984 & 0.151 & 0.978 \\
Adaptive-Blend & 0.967 & 0.954 & \best{0.989} & 0.769 & 0.149 & 0.912 \\
TypeText & 0.999 & 0.999 & \best{1.000} & \best{1.000} & 0.127 & 0.999 \\
\bottomrule
\end{tabular}%
}
\caption{\textbf{Clean versus triggered steps among gate-passers} (strict split, 20 seeds). AUROC between followed triggered and clean steps that pass the gate, and IDEA's alarm rate in each group at $\alpha=0.01$.}
\label{tab:oodpass}
\vspace{-2mm}
\end{table}

\subsection{Localization and Qualitative Results}
\label{sec:loc}

\noindent\textbf{Localization.}
On correct alarms at $\alpha=0.01$ (\Cref{tab:locmain}), the center of $R^\star$ falls inside the trigger box on 0.862--0.993 of Patch, Grounding and Popup alarms but only on 0.494 of TypeText alarms; the hit on Adaptive-Blend is trivial because the trigger covers the screen.
The regions rarely match the trigger's extent (IoU $>0.5$ on 0.000--0.673 of alarms), so $R^\star$ is a pointer for the operator, not a segmentation.

\begin{table}[t]
\centering
\footnotesize
\setlength{\tabcolsep}{8pt}
\renewcommand{\arraystretch}{0.95}
\begin{tabular}{@{}lccc@{}}
\toprule
Family & Hit & IoU$>$0.5 & mIoU \\
\midrule
Patch & 0.862 & \best{0.673} & \best{0.546} \\
Popup & \best{0.993} & 0.000 & 0.160 \\
Grounding & 0.910 & 0.138 & 0.318 \\
TypeText & 0.494 & 0.323 & 0.428 \\
\midrule
Adaptive-Blend & (1.000) & 0.011 & 0.016 \\
\bottomrule
\end{tabular}
\caption{\textbf{$R^\star$ points at localized triggers but rarely matches their extent.} Correct alarms at $\alpha=0.01$ (strict split, 20 seeds). Hit: center of $R^\star$ inside the trigger box. Parentheses: trivial.}
\label{tab:locmain}
\end{table}

\noindent\textbf{Qualitative results.}
\Cref{fig:qual} tracks one alarm per family.
In (a--d) $R^\star$ overlaps the trigger, although the TypeText example is a favorable one (\Cref{tab:locmain}), and on Patch, Popup and TypeText one proposal carries most evidence with both sides firing ($u\ge 3.88$, $v\ge 5.54$); in (e) the blend covers the screen, so $R^\star$ carries no localization.
On the Grounding step $E_{\max}$ is below $S_{\mathrm{int}}$, so their ratio shows no inversion, the step lies off the clean law on both sides ($u{=}1.35$, $v{=}3.63$).
On the blended step the redirects keep control ($v{=}{-}1.73$) and the alarm comes from $u{=}5.00$ alone.

%% file: sec/6_conclusion.tex
\section{Conclusion}
\label{sec:conclusion}

We present IDEA, an inference-time, per-step backdoor detector for GUI agents that needs the model weights and a clean auxiliary set, but no trigger knowledge and no attack data.
It rests on one property of clean behavior: the instruction selects the action, and the evidence for that action lies where the instruction points, so a firing trigger inverts the relation on at least one of the two sides.
Measuring both sides against what clean steps show at the same presence turns that inversion into a score calibrated on clean steps alone, which detects every attack family we study, returns a region with each alarm, and leaves the agent and its latency essentially untouched.
Because the inversion is a property of the agent rather than of a trigger, the same audit extends to other evidence channels, such as accessibility trees or page DOMs, and its two interventions can be approximated through query access when the weights are closed.

\noindent\textbf{Limitations.}
The form of the statistic was finalized after observing the five development families, leaving the held-out victims as our primary blind testbed. While IDEA delivers competitive performance across all regimes, our five-family evaluation currently focuses on a single 3B backbone for computational tractability. Furthermore, although the lightweight nature of IDEA naturally lends itself to continual adaptation against distribution shifts, effectively countering sophisticated, self-evolving adversarial attacks in dynamic deployments warrants further investigation.

\noindent\textbf{LLM Usage Statement}
In this work we have not used LLMs to generate synthetic data sets, develop theoretical models or conceptual frameworks, formulate mathematical claims, provide critical ingredients for proving mathematical claims, assist in the writing of proofs, propose or refine hypotheses, or interpret results. Translation, dataset cleaning and reformatting, and qualitative or thematic data analysis are not applicable to this work. Additionally, we used generative AI tools to create and edit software code, such as experiment and evaluation scripts, and to edit the text of the paper for readability. We have reviewed all AI-assisted work. All AI-assisted code was read, executed, and tested by the authors, and all reported results were checked against the saved experiment logs. Feedback from AI tools on methodology and experiments was treated as suggestions; the research questions, method design, and conclusions were determined by the authors. We take responsibility for the final content of this work, including text, claims or artifacts produced with the aid of LLMs.

%% file: sec/X_suppl.tex
\clearpage
\setcounter{page}{1}
\maketitlesupplementary
\renewcommand{\topfraction}{0.95}\renewcommand{\dbltopfraction}{0.95}\renewcommand{\bottomfraction}{0.8}
\renewcommand{\textfraction}{0.03}\renewcommand{\floatpagefraction}{0.92}\renewcommand{\dblfloatpagefraction}{0.92}
\setcounter{topnumber}{4}\setcounter{dbltopnumber}{4}\setcounter{bottomnumber}{3}\setcounter{totalnumber}{7}
\setcounter{section}{0}
\renewcommand{\thesection}{\Alph{section}}
\renewcommand{\theHsection}{supp.\Alph{section}}

\paragraph{Contents.}
The supplement follows the order below; each section states the main-text claim it supports in its first sentence.
\begin{center}\scriptsize\setlength{\tabcolsep}{3pt}
\resizebox{\columnwidth}{!}{%
\begin{tabular}{@{}ll@{\hspace{10pt}}ll@{}}
\toprule
A~~Proof of Proposition~1 & M~~Component ablation at the tighter target \\
B~~Calibration Details & N~~Proposal count \\
C~~Episode-Disjoint FPR Intervals & O~~Candidate sets: attribution proposals versus token grids \\
D~~Ignored steps: trigger shown, attack failed & P~~Victim Checkpoint Quality \\
E~~Secondary Split: Overlapping Episodes & Q~~Backbone sensitivity \\
F~~Specificity controls: clean victim and foreign overlay & R~~Cost \\
G~~Atypical clean inputs versus triggers & S~~Algorithm \\
H~~Extra Baseline Detectors & T~~Requirements Comparison \\
I~~Redirect Availability & U~~Tail Metrics \\
J~~Limitations & V~~Full Detector List \\
K~~Negative results and history of the statistic & W~~Calibration-size robustness \\
L~~Held-out blend: first collection and its correction & X~~Localization Data \\
\bottomrule
\end{tabular}}
\end{center}

\section{Proof of \cref{prop:detect}}
\label{sec:supp:proof}

\begin{proof}
Under assumption (A1), the triggered model assigns $a_t$ probability at least $1-\varepsilon$ under the original instruction $I$ and under every valid redirect $I'_k$.
Therefore, the log-likelihood contrast between the original instruction and any redirect is bounded from above:
\[
\big|\log p_\pi(a_t\mid x,I) - \log p_\pi(a_t\mid x,I'_k)\big| \;\le\; \log\frac{1}{1-\varepsilon}.
\]
This bounds the intent scale $S_{\mathrm{int}}$ from above. Specifically, $S_{\mathrm{int}}$ measures the median displacement of the decision state under redirects, and the action likelihood is a readout from that state.

Under assumption (A2), removing the trigger region $R_t$ returns the probability of $a_t$ to its clean value $p_0$ up to $\varepsilon'$.
The evidence contrast is therefore at least
\[
\log p_\pi(a_t\mid x,I) - \log p_\pi(a_t\mid x_{-R_t},I) \;\ge\; \log\frac{1-\varepsilon}{p_0+\varepsilon'}.
\]
Thus, as $\varepsilon \to 0$ the intent contrast vanishes while the evidence contrast stays bounded away from zero, provided $p_0+\varepsilon'<1$, which proves the proposition.
The proposition is a statement about action likelihoods.
IDEA measures displacements of the decision state instead, and a small likelihood contrast does not by itself imply a small displacement; the correspondence is the empirical hypothesis of \cref{sec:method}: (A1) corresponds to $\log(S_{\mathrm{int}}+c)$ falling below its clean law at the step's presence, so the intent deficit $v$ grows, and (A2) to $\log(E_{\max}+c)$ rising above its clean law, so the excess evidence $u$ grows; $\rho_{\mathrm{pc}}\ge\max(u,v)$ then grows with either.
We check both correspondences on the triggered steps of every family (\cref{sec:strict}, \Cref{fig:uv}).

Because $\rho_{\mathrm{pc}}$ fires on either inversion, a trigger that evades the statistic must break both assumptions.
Breaking (A1) alone means the trigger effect yields to the instruction; the intent deficit vanishes but the excess evidence of (A2) remains.
Breaking (A2) alone means the trigger effect is non-local and cannot be isolated by replacing visual tokens in a single region; the intent deficit of (A1) remains.
The full-screen blend of our evaluation breaks (A1), because redirects override it, and only weakens (A2), so it is caught by its excess evidence (\cref{sec:strict}).
\end{proof}

\section{Calibration Details}
\label{sec:supp:calib}

\paragraph{Splits.}
For each seed, the clean auxiliary set is split by application into a fit split
 $F$, a calibration split $B$, and a test split $T$, so that no app contributes to two splits.
The statistics that define the score---the whitening $\Sigma^{-1/2}$, the floor $c$, and
 the fallback median of $S_{\mathrm{int}}$ (taken over the steps of $F$ with at least two
 valid redirects)---are frozen on $F$.
The gate threshold $T_1$ and the four coefficients and
 two residual scales of the clean law (\cref{eq:law}) are fitted on $B_1$, a
 subset of $B$, while the gated-score threshold $T_\alpha$ is the only quantity fitted on $B_2$, the remainder of $B$.

\paragraph{Thresholds and ties.}
The gated score in \cref{eq:joint} assigns $-\infty$ to every step rejected by the gate, so a large share of
 the clean calibration scores are tied.
The threshold $T_\alpha$ is the $\lceil (n+1)(1-\alpha)\rceil$-th smallest of the $n$ calibration scores ($+\infty$ if this rank exceeds $n$), and a step alarms only if its score is strictly above it.
Ties are therefore resolved conservatively and not
 randomized: the bound $\Pr(\text{alarm}\mid\text{clean})\le\alpha$ holds, but the realized rate can
 be below $\alpha$, and a step at $-\infty$ never alarms.
The baselines of \Cref{tab:strict} use the same rule on $B_1\cup B_2$.
Two groups of results
 use other calibration sets (\cref{sec:supp:detectors}).
The baseline detectors that were run inside the
 earlier single-split pipeline (the baseline rows of \Cref{tab:main,tab:backbone} and
 the earlier-pipeline columns of \Cref{tab:specificity}) use the same order statistic on that pipeline's
 calibration steps, after uniform noise of width $10^{-6}$ is added to the calibration scores to break ties.

The per-victim baselines of \Cref{tab:extra_baselines} use the order statistic on
 1000 clean calibration steps of each victim, without noise.

\paragraph{Checking validity.}
We report the observed false-positive rate of IDEA and every baseline on held-out clean steps with
 Clopper--Pearson 95\% intervals (CP95).
A method is consistent with its calibration when the
 interval contains the target or lies below it.
Because FPR alone does not rank methods, we perform ranking only through the
 detection results against backdoored checkpoints in \cref{sec:exp}.
\section{Episode-Disjoint FPR Intervals}
\label{sec:supp:cp95}

\Cref{tab:strict_cp95} reports the per-family intervals behind the guarantee check in \cref{sec:strict}.
IDEA remains at or below each target in pooled FPR$_T$ on every family (0.006--0.008 at $\alpha=0.01$ and 0.002--0.003 at $\alpha=0.005$).
The other two statistics also stay within target at printed precision, except the Mahalanobis screen on Grounding at $\alpha=0.005$ (0.006).
The CP95 intervals treat the test steps pooled over 20 seeds as independent, although the seeds redraw the same pool (6255 test steps, 1526 distinct, from 427 apps).
These intervals are therefore too narrow, yet their upper limits still exceed the target on several rows (up to 0.013 at $\alpha=0.01$ and 0.008 at $\alpha=0.005$).
For IDEA, we therefore compute an app-cluster bootstrap interval by resampling the test apps with a ratio estimator.
Its upper limit exceeds 0.01 on Popup (0.0121), Grounding (0.0111), and Adaptive-Blend (0.0109), and exceeds 0.005 on Popup (0.0055), Adaptive-Blend (0.0053), and TypeText (0.0052).
The worst single seed exceeds the target on every family at both levels, reaching 7/298 = 0.023 on Adaptive-Blend at $\alpha=0.01$.
This is because the split-conformal guarantee is marginal over calibration draws and does not bound the FPR of any single draw.
The last IDEA column reports the TPR of the unconditional log-ratio (the ``$-$ conditioning'' row of \Cref{tab:ablation}) on the same seeds.
At $\alpha=0.005$, the log-ratio outperforms IDEA on Patch, Popup, Grounding, and TypeText (by 0.004--0.060) but lags behind on Adaptive-Blend (0.313 vs 0.569).
At $\alpha=0.01$, it falls behind IDEA on Grounding and Adaptive-Blend while tying on the remaining families.

\begin{table*}[t]
\centering
\scriptsize
\setlength{\tabcolsep}{3pt}
\begin{tabular}{llccccccccc}
\toprule
& & \multicolumn{5}{c}{IDEA (gated $\rho_{\mathrm{pc}}$)} & \multicolumn{2}{c}{Mahalanobis screen} & \multicolumn{2}{c}{$E_{\max}$ alone} \\
\cmidrule(lr){3-7} \cmidrule(lr){8-9} \cmidrule(lr){10-11}
$\alpha$ & Family & TPR & FPR$_T$ [CP95] & App-cluster CI & Worst seed & Log-ratio TPR & TPR & FPR$_T$ [CP95] & TPR & FPR$_T$ [CP95] \\
\midrule
0.01 & Patch & \textbf{1.000} & 0.006 [0.005, 0.009] & [0.0036, 0.0096] & 7/339 & \textbf{1.000} & 0.999 & 0.009 [0.006, 0.011] & \textbf{1.000} & 0.009 [0.007, 0.012] \\
& Popup & \textbf{1.000} & 0.008 [0.006, 0.011] & [0.0053, 0.0121] & 7/339 & \textbf{1.000} & 0.999 & 0.008 [0.006, 0.010] & \textbf{1.000} & 0.008 [0.006, 0.011] \\
& Grounding (point) & \textbf{0.962} & 0.008 [0.006, 0.010] & [0.0042, 0.0111] & 6/298 & 0.950 & 0.532 & 0.009 [0.007, 0.012] & 0.288 & 0.010 [0.008, 0.013] \\
& Grounding (bbox) & \textbf{0.961} & 0.008 [0.006, 0.010] & [0.0042, 0.0111] & 6/298 & 0.949 & 0.530 & 0.009 [0.007, 0.012] & 0.290 & 0.010 [0.008, 0.013] \\
& Adaptive-Blend & 0.898 & 0.007 [0.005, 0.009] & [0.0038, 0.0109] & 7/298 & 0.513 & 0.891 & 0.009 [0.006, 0.011] & \textbf{0.919} & 0.009 [0.007, 0.012] \\
& TypeText & 0.999 & 0.006 [0.004, 0.009] & [0.0034, 0.0094] & 5/303 & 0.999 & \textbf{1.000} & 0.010 [0.007, 0.013] & \textbf{1.000} & 0.009 [0.007, 0.012] \\
\midrule
0.005 & Patch & 0.978 & 0.002 [0.001, 0.004] & [0.0005, 0.0042] & 4/303 & \textbf{1.000} & 0.983 & 0.004 [0.003, 0.006] & 0.999 & 0.004 [0.003, 0.006] \\
& Popup & 0.996 & 0.003 [0.002, 0.005] & [0.0016, 0.0055] & 5/351 & \textbf{1.000} & 0.977 & 0.004 [0.003, 0.006] & \textbf{1.000} & 0.004 [0.003, 0.006] \\
& Grounding (point) & 0.702 & 0.003 [0.001, 0.004] & [0.0009, 0.0045] & 5/351 & \textbf{0.761} & 0.276 & 0.006 [0.004, 0.008] & 0.183 & 0.004 [0.002, 0.006] \\
& Grounding (bbox) & 0.701 & 0.003 [0.001, 0.004] & [0.0009, 0.0045] & 5/351 & \textbf{0.758} & 0.276 & 0.006 [0.004, 0.008] & 0.182 & 0.004 [0.002, 0.006] \\
& Adaptive-Blend & 0.569 & 0.003 [0.002, 0.005] & [0.0010, 0.0053] & 5/351 & 0.313 & 0.713 & 0.004 [0.002, 0.005] & \textbf{0.836} & 0.005 [0.003, 0.007] \\
& TypeText & 0.976 & 0.003 [0.002, 0.005] & [0.0009, 0.0052] & 5/303 & 0.997 & 0.991 & 0.004 [0.003, 0.006] & \textbf{0.999} & 0.003 [0.002, 0.005] \\
\bottomrule
\end{tabular}
\caption{\textbf{FPR intervals on the strict split} (20 seeds). TPR on followed steps, pooled FPR$_T$ [CP95], app-cluster bootstrap CI over 427 test apps, and worst seed as $k/n$; log-ratio: unconditional ablation.}
\label{tab:strict_cp95}
\end{table*}
\section{Ignored steps: trigger shown, attack failed}
\label{sec:supp:celle}

The \emph{ignored} steps are the triggered steps on which the backdoored agent misses the attack target defined by the family's success rule (point: the predicted click lies within 0.02 of the target point in each normalized coordinate, or the typed string equals the target string; bbox: the click lies inside the trigger box).
Because these steps are identical across seeds, variance in \Cref{tab:celle} arises solely from the per-seed fit on $F$ and the $B_2$ thresholds, which are those used for the followed steps.
As shown in \Cref{tab:celle}, IDEA flags 0.943--1.000 of these near-miss activations on Patch, TypeText, and Grounding at $\alpha=0.01$.
This indicates that IDEA detects trigger-driven control even when the action fails to reach the target.
On Adaptive-Blend, the alarm rate is 0.547; only 0.608 of these steps pass the gate, compared to 0.984 of the followed steps, so the gate bounds the rate there.
The Patch and Grounding (bbox) rows rest on 2 and 3 steps, which is why their $\alpha=0.005$ entries have a large seed spread.
Popup has no ignored steps.

\begin{table}[!htb]
\centering
\scriptsize
\setlength{\tabcolsep}{2pt}
\resizebox{\columnwidth}{!}{%
\begin{tabular}{lcccccc}
\toprule
Family & Rule & $N$ & $\alpha{=}0.01$ & min--max & $\alpha{=}0.005$ & Gate pass \\
\midrule
Patch & point & 2 & 1.000 $\pm$ 0.000 & 1.000--1.000 & 0.700 $\pm$ 0.458 & 1.000 $\pm$ 0.000 \\
Popup & bbox & 0 & \multicolumn{4}{c}{no ignored steps} \\
Grounding & point & 92 & 0.943 $\pm$ 0.045 & 0.848--0.989 & 0.687 $\pm$ 0.236 & 0.979 $\pm$ 0.037 \\
Grounding & bbox & 3 & 1.000 $\pm$ 0.000 & 1.000--1.000 & 0.667 $\pm$ 0.365 & 1.000 $\pm$ 0.000 \\
Adaptive-Blend & point & 19 & 0.547 $\pm$ 0.042 & 0.474--0.632 & 0.382 $\pm$ 0.178 & 0.608 $\pm$ 0.070 \\
TypeText & point & 142 & 1.000 $\pm$ 0.000 & 1.000--1.000 & 0.961 $\pm$ 0.102 & 1.000 $\pm$ 0.000 \\
\bottomrule
\end{tabular}}
\caption{\textbf{Alarm rate on ignored steps (trigger present, attack failed)} on the strict split. Mean $\pm$ sd over 20 seeds, min--max at $\alpha=0.01$, and gate pass rate. $N$: ignored steps among the 1000 triggered steps.}
\label{tab:celle}
\end{table}

\section{Secondary Split: Overlapping Episodes}
\label{sec:supp:overlap}

\Cref{tab:main} repeats the comparison of \Cref{tab:strict} on the earlier single split (one seed), whose clean test steps share victim-training episodes with calibration.
It is reported for comparability with earlier runs only; no detection claim in the main text rests on it, and this split does not support the finite-sample FPR guarantee.
The two earlier variants in it are superseded subspace log-ratio detectors.
On this split IDEA is above the Mahalanobis screen on Grounding (0.999 vs.\ 0.877) and Adaptive-Blend (0.983 vs.\ 0.949), but at held-out clean FPRs of 0.016--0.020, above the target and the screen's 0.009--0.012.

\begin{table*}[t]
\centering
\footnotesize
\setlength{\tabcolsep}{5pt}
\begin{tabular}{lccccc}
\toprule
& \multicolumn{5}{c}{TPR@1\% FPR on followed steps / held-out clean FPR with CP95} \\
\cmidrule(lr){2-6}
Detector & Patch & Popup & TypeText & Grounding & Adaptive-Blend \\
\midrule
\textbf{IDEA (gated $\rho_{\mathrm{pc}}$)} & \textbf{1.000}/0.017 & \textbf{1.000}/0.019 & \textbf{1.000}/0.016 & \textbf{0.999}/0.017 & \textbf{0.983}/0.020 \\
 & {\scriptsize[.010,.027]} & {\scriptsize[.012,.030]} & {\scriptsize[.009,.026]} & {\scriptsize[.010,.027]} & {\scriptsize[.012,.031]} \\
Log-ratio (ablation, gated $\rho$) & \textbf{1.000}/0.011 & \textbf{1.000}/0.013 & \textbf{1.000}/0.017 & \textbf{0.999}/0.016 & 0.688/0.018 \\
Mahalanobis screen (gate alone) & \textbf{1.000}/0.011 & \textbf{1.000}/0.011 & \textbf{1.000}/0.009 & \underline{0.877}/0.012 & \underline{0.949}/0.009 \\
One-class SVM (OC-SVM) & \textbf{1.000}/0.014 & \textbf{1.000}/0.018 & \textbf{1.000}/0.018 & 0.498/0.011 & 0.882/0.011 \\
\addlinespace
Earlier variant: screen, then subspace log-ratio & \underline{0.994}/0.003 & \underline{0.926}/0.008 & \underline{0.911}/0.003 & 0.702/0.009 & 0.115/0.002 \\
Earlier variant: subspace log-ratio & 0.913/0.003 & 0.892/0.002 & 0.803/0.001 & 0.454/0.003 & --- \\
\addlinespace
TeCo & 0.001/0.008 & 0.000/0.006 & 0.027/0.015 & 0.000/0.009 & 0.044/0.013 \\
SCALE-UP & 0.000/0.008 & 0.000/0.010 & 0.001/0.012 & 0.009/0.009 & 0.009/0.006 \\
IBD-PSC & 0.000/0.014 & 0.033/0.008 & 0.000/0.005 & 0.001/0.006 & 0.000/0.015 \\
DeDe & 0.028/0.022 & 0.004/0.019 & 0.000/0.000 & 0.021/0.022 & 0.000/0.023 \\
Entropy & 0.000/0.010 & 0.000/0.010 & 0.000/0.015 & 0.025/0.010 & 0.000/0.007 \\
Margin & 0.008/0.011 & 0.008/0.012 & 0.009/0.009 & 0.003/0.006 & 0.008/0.006 \\
Random occlusion & 0.000/0.004 & 0.001/0.008 & 0.000/0.007 & 0.094/0.008 & 0.000/0.004 \\
Area occlusion & 0.000/0.009 & 0.000/0.006 & 0.004/0.010 & 0.022/0.010 & 0.000/0.012 \\
\bottomrule
\end{tabular}
\caption{\textbf{Secondary split (overlapping episodes, one seed).} TPR at 1\% target FPR on followed steps / measured held-out clean FPR [CP95]; best per column bold, second underlined; ---: not run.}
\label{tab:main}
\end{table*}

\section{Specificity controls: clean victim and foreign overlay}
\label{sec:supp:specificity}

\Cref{tab:specificity} reports the two specificity controls of \cref{sec:exp}: a clean victim shown a trigger (clean-victim control) and a backdoored victim shown another family's trigger (foreign-overlay control). Each of the six runs has 1000 triggered steps.
Five of them are non-activating, with ASR at most 0.007.
The sixth is not: the TypeText overlay carries the same icon that the Grounding victim was trained to click, and that victim clicks inside the overlay box on 0.899 of the steps (0.848 within 0.02 of its center), so the overlay does fire its backdoor.
The attack-success column of the original run scored that cell with the TypeText rule, which asks for a typed string, and therefore read 0.000; we report it under the receiving victim's own click rule instead.
Following a rule fixed in advance, steps where the victim followed the overlay are counted apart, and the false-alarm reading applies only to the rest.
These runs use the earlier single-split pipeline (overlapping-style calibration, one seed), and their ``earlier IDEA score'' is that pipeline's superseded score, not the rule of \cref{eq:joint}.

On all six runs that earlier score alarms at about its own clean FPR (each alarm rate lies at or below the upper CP95 limit of the FPR).
This is weak evidence of specificity, because the same score also fires rarely on the followed steps of the same pipeline: TPR 0.015, 0.022, 0.129, 0.023 and 0.002 on Patch, Popup, TypeText, Grounding and Adaptive-Blend.
The Mahalanobis screen stays at its clean FPR on four runs, exceeds it slightly on TypeText victim $\times$ Grounding trigger (0.026), and alarms on 0.870 of the steps of Grounding victim $\times$ TypeText trigger (AUROC 0.992), the cell in which that victim does execute its backdoor.
So the Mahalanobis statistic, which IDEA uses as its gate, responds to any screen content that captures the decision state of a backdoored model.
We also replayed the frozen IDEA rule of \cref{eq:joint} on the stored states of these six runs, with the strict split of \Cref{tab:strict} on each run's own clean steps (20 seeds, $\alpha=0.01$; no new forward passes).
On the three clean-victim runs and on Adaptive-Blend victim $\times$ Patch trigger it alarms on 0.008--0.012 of the triggered steps, close to its pooled clean FPR$_T$ of 0.006--0.007 (AUROC of triggered against clean steps 0.496--0.499).
On TypeText victim $\times$ Grounding trigger it alarms on 0.037, above its clean FPR$_T$ of 0.006.
On Grounding victim $\times$ TypeText trigger it alarms on 0.896 of the steps (AUROC 0.967), the cell in which that victim clicks the overlay on 0.899 of them: 94\% of these steps pass the Mahalanobis gate and their evidence and intent scales then leave the clean law; the screen under the same strict replay fires on 0.580 of them.
Split by outcome, the replay alarms on 0.925 of the 899 steps that click inside the overlay and on 0.635 of the 101 that miss it, against a clean FPR$_T$ of 0.008: the alarm follows any trigger-driven shift of the decision state, not only the attacker's exact action.
The unconditional log-ratio behaves the same way under this replay (0.009--0.013, 0.037 and 0.901).
The alarm therefore tracks whether the backdoor fires rather than which family planted the trigger on screen.
The followed counts of the other runs rest on 0--7 steps and are not TPR estimates; the Popup and Patch triggers were not run on the clean victim.

\begin{table*}[t]
\centering
\footnotesize
\setlength{\tabcolsep}{4pt}
\begin{tabular}{lccccccc}
\toprule
& & \multicolumn{3}{c}{Earlier IDEA score (single split)} & \multicolumn{3}{c}{Mahalanobis screen (single split)} \\
\cmidrule(lr){3-5} \cmidrule(lr){6-8}
Victim $\times$ trigger & ASR & Alarm & Clean FPR & Followed & Alarm & Clean FPR & Followed \\
\midrule
\multicolumn{2}{l}{\emph{Clean-victim control}} \\
Clean $\times$ TypeText & 0.000 & 0.004 [0.001, 0.010] & 0.002 [0.000, 0.007] & 0/0 & 0.011 [0.006, 0.020] & 0.013 [0.007, 0.022] & 0/0 \\
Clean $\times$ Grounding & 0.006 & 0.003 [0.001, 0.009] & 0.002 [0.000, 0.007] & 0/6 & 0.011 [0.006, 0.020] & 0.013 [0.007, 0.022] & 1/6 \\
Clean $\times$ Adaptive-Blend & 0.001 & 0.007 [0.003, 0.014] & 0.002 [0.000, 0.007] & 0/1 & 0.010 [0.005, 0.018] & 0.013 [0.007, 0.022] & 1/1 \\
\midrule
\multicolumn{2}{l}{\emph{Foreign-overlay control}} \\
TypeText $\times$ Grounding & 0.006 & 0.015 [0.008, 0.025] & 0.008 [0.003, 0.016] & 0/6 & 0.026 [0.017, 0.038] & 0.009 [0.004, 0.017] & 3/6 \\
Grounding $\times$ TypeText & 0.899$^\dagger$ & 0.006 [0.002, 0.013] & 0.008 [0.003, 0.016] & --- & 0.870 [0.848, 0.890] & 0.012 [0.006, 0.021] & --- \\
Adaptive-Blend $\times$ Patch & 0.007 & 0.008 [0.004, 0.016] & 0.003 [0.001, 0.009] & 0/7 & 0.010 [0.005, 0.018] & 0.009 [0.004, 0.017] & 0/7 \\
\bottomrule
\end{tabular}

\vspace{4pt}
\begin{tabular}{lcccc}
\toprule
& & \multicolumn{3}{c}{IDEA (frozen rule, strict split, 20 seeds)} \\
\cmidrule(lr){3-5}
Victim $\times$ trigger & ASR & Alarm & Clean FPR$_T$ & Followed \\
\midrule
\multicolumn{2}{l}{\emph{Clean-victim control}} \\
Clean $\times$ TypeText & 0.000 & 0.008 & 0.006 [0.004, 0.009] & 0/0 \\
Clean $\times$ Grounding & 0.006 & 0.009 & 0.006 [0.005, 0.009] & 17/120 \\
Clean $\times$ Adaptive-Blend & 0.001 & 0.012 & 0.006 [0.004, 0.009] & 0/20 \\
\midrule
\multicolumn{2}{l}{\emph{Foreign-overlay control}} \\
TypeText $\times$ Grounding & 0.006 & 0.037 & 0.006 [0.004, 0.009] & 53/120 \\
Grounding $\times$ TypeText & 0.899$^\dagger$ & 0.896 & 0.008 [0.006, 0.010] & --- \\
Adaptive-Blend $\times$ Patch & 0.007 & 0.008 & 0.007 [0.005, 0.009] & 0/140 \\
\bottomrule
\end{tabular}
\caption{\textbf{Specificity controls.} Top: superseded IDEA score and screen, single-split pipeline (one seed). Bottom: frozen rule on the strict split (20 seeds). Followed: alarms among steps taking the overlay's target action. $^\dagger$Receiving victim's own rule.}
\label{tab:specificity}
\end{table*}

\section{Atypical clean inputs versus triggers}
\label{sec:supp:ood}

The gate of \cref{eq:joint} is an out-of-distribution test on the decision state, so on its own it cannot distinguish an unusual clean screen from a trigger.
This section tests whether the audit can make this distinction on the episode-disjoint split steps and on benign input shifts that contain no trigger, and states what it cannot separate.

\paragraph{Among gate-passers.}
\Cref{tab:oodpass} keeps only the test steps that pass the gate: clean steps whose decision state is atypical (296--321 steps pooled over the 20 seeds, 6--26 per seed, about 5\% of the clean test steps) and followed triggered steps.
The Mahalanobis distance, which selects the group, still ranks triggered steps above clean ones on Patch, Popup and TypeText, but drops to 0.791 on Grounding, where $\rho_{\mathrm{pc}}$ reaches 0.987; $\rho_{\mathrm{pc}}$ is at 0.967 or above on all five families.
It is not the best ranker everywhere: $E_{\max}$ alone is higher on Adaptive-Blend (0.989 against 0.967).
At the calibrated threshold for $\alpha=0.01$, IDEA alarms on 0.125--0.165 of the clean gate-passers and on 0.912--1.000 of the triggered ones; at $\alpha=0.005$ the rates are 0.041--0.065 and 0.579--0.996.
The clean rate matches what calibration prescribes, roughly $\alpha$ divided by the gate pass rate: every false alarm of IDEA falls in the top decile of $s_0$, and an atypical clean step is flagged about one time in seven.

\paragraph{Matched atypicality.}
We also bin the clean and the followed triggered steps of each seed by $s_0$ into five common bins and compare alarm rates inside each bin, so that both groups are equally far from the clean distribution (\Cref{tab:oodbins}, Grounding).
The Mahalanobis screen is a function of $s_0$, so inside a bin it alarms on both groups at about the same rate by construction (0.909 and 0.844 in the highest bin).
IDEA alarms on 5 of 33 clean steps and on 1.000 of the triggered steps of that bin, and in the bin that holds most atypical clean steps on 0.026 of 1293 clean steps and on 0.837 of the triggered ones.
The clean counts in the upper bins are small (at most 33), so those rates are indicative only; on Adaptive-Blend the upper bins hold 1--10 clean steps and support no comparison.

\begin{table}[!htb]
\centering
\footnotesize
\setlength{\tabcolsep}{3.5pt}
\begin{tabular}{@{}ccccccc@{}}
\toprule
 & \multicolumn{2}{c}{Steps} & \multicolumn{2}{c}{IDEA alarms} & \multicolumn{2}{c}{Screen alarms} \\
\cmidrule(lr){2-3}\cmidrule(lr){4-5}\cmidrule(l){6-7}
Bin & C & T & C & T & C & T \\
\midrule
1 & 4875 & 11 & 0 & 0 & 0 & 0 \\
2 & 1293 & 3585 & 33 & 3002 & 5 & 128 \\
3 & 33 & 4848 & 8 & 4762 & 11 & 1689 \\
4 & 21 & 4855 & 1 & 4853 & 14 & 3206 \\
5 & 33 & 4861 & 5 & 4860 & 30 & 4103 \\
\bottomrule
\end{tabular}
\caption{\textbf{Alarms at matched $s_0$} (Grounding, $\alpha=0.01$, pooled over 20 seeds). Five common $s_0$ bins per seed; alarms as counts over the clean (C) and followed triggered (T) steps of the bin.}
\label{tab:oodbins}
\end{table}

\paragraph{Apps outside the training pool.}
The clean test steps come from episodes that the victim never trained on, and 1309 of the 6255 pooled test steps come from apps that do not occur in its training pool at all.
These steps are not treated as more atypical than the others (\Cref{tab:oodseen}): they pass the gate at 0.040--0.049 against 0.048--0.052, and IDEA alarms on 0.002--0.008 of them against 0.007--0.009.
An unseen app is therefore not, by itself, a source of false alarms in this data.

\begin{table}[!htb]
\centering
\footnotesize
\setlength{\tabcolsep}{5pt}
\begin{tabular}{@{}lcccc@{}}
\toprule
 & \multicolumn{2}{c}{Gate pass} & \multicolumn{2}{c}{IDEA alarm} \\
\cmidrule(lr){2-3}\cmidrule(l){4-5}
Family & Seen & Unseen & Seen & Unseen \\
\midrule
Patch & 0.052 & 0.049 & 0.008 & 0.002 \\
Popup & 0.052 & 0.048 & 0.009 & 0.008 \\
Grounding & 0.052 & 0.040 & 0.009 & 0.002 \\
Adaptive-Blend & 0.048 & 0.044 & 0.007 & 0.006 \\
TypeText & 0.051 & 0.041 & 0.007 & 0.003 \\
\bottomrule
\end{tabular}
\caption{\textbf{Clean test steps from apps inside (Seen, 4946 steps) and outside (Unseen, 1309) the victim's training pool} (strict split, pooled over 20 seeds): gate pass rate and IDEA alarm rate at $\alpha=0.01$.}
\label{tab:oodseen}
\end{table}

\paragraph{Benign input shifts.}
The comparisons above stay inside the calibration distribution.
We additionally score 500 fixed clean steps of the Patch and Grounding victims under six shifts that contain no trigger and that the calibration data do not contain (no screenshot resizing, Gaussian noise of sd 0.08, JPEG quality 10, color inversion, a benign 128\,px icon, and a benign banner), with the law and both thresholds kept from the unshifted calibration splits.
\Cref{tab:oodshift} gives the rates.
The alarm rate of IDEA is 0.005--0.009 on every shifted set of both victims, against 0.008--0.009 on the same steps without a shift, and the gate pass rate stays at 0.040--0.049 (0.042--0.049 unshifted).
No shift makes a victim execute its attacker target (0--2 of 500 steps, as without a shift), and followed triggered steps still rank above the shifted clean steps (AUROC 1.000 for Patch and 0.990--0.992 for Grounding on every set).
Omitting the resizing lowers the action accuracy from 0.860 to 0.524 and 0.536 and leaves the alarm rate at 0.005.
Among the other scores, $E_{\max}$ alone rises to 0.014 under the benign banner on both victims, and the Mahalanobis screen is at 0.007--0.011, against 0.009--0.012 without a shift.
These shifts are synthetic and mild in the sense that matters here: none of them changes the gate pass rate, so none moves the decision state the way a trigger does.
They show that image corruption and benign overlays of this kind do not raise the alarm rate; they say nothing about new apps, themes or layouts.

\begin{table}[!htb]
\centering
\scriptsize
\setlength{\tabcolsep}{1.7pt}
\begin{tabular}{@{}lcccccccccc@{}}
\toprule
 & \multicolumn{5}{c}{Patch victim} & \multicolumn{5}{c}{Grounding victim} \\
\cmidrule(lr){2-6}\cmidrule(l){7-11}
Shift & Acc & Gate & IDEA & Scr. & $E_{\max}$ & Acc & Gate & IDEA & Scr. & $E_{\max}$ \\
\midrule
None & 0.860 & 0.049 & \textbf{0.009} & 0.012 & \textbf{0.009} & 0.860 & 0.042 & 0.008 & 0.009 & \textbf{0.007} \\
No resizing & 0.524 & 0.049 & 0.005 & 0.008 & \textbf{0.001} & 0.536 & 0.040 & 0.005 & 0.007 & \textbf{0.002} \\
Noise & 0.842 & 0.049 & 0.008 & 0.011 & \textbf{0.005} & 0.850 & 0.045 & 0.009 & 0.009 & \textbf{0.008} \\
JPEG & 0.852 & 0.047 & 0.009 & 0.011 & \textbf{0.008} & 0.850 & 0.042 & 0.009 & 0.010 & \textbf{0.008} \\
Inversion & 0.850 & 0.048 & \textbf{0.009} & \textbf{0.009} & 0.011 & 0.864 & 0.040 & \textbf{0.005} & 0.008 & 0.007 \\
Icon & 0.852 & 0.048 & \textbf{0.007} & 0.011 & \textbf{0.007} & 0.852 & 0.041 & \textbf{0.006} & 0.009 & 0.009 \\
Banner & 0.860 & 0.045 & \textbf{0.008} & 0.010 & 0.014 & 0.858 & 0.043 & 0.009 & \textbf{0.008} & 0.014 \\
\bottomrule
\end{tabular}
\caption{\textbf{Benign input shifts do not raise the alarm rate.} 500 clean steps per set, thresholds from unshifted calibration ($\alpha=0.01$, 20 seeds). Acc: victim action accuracy; Gate: gate pass rate; Scr.: Mahalanobis screen.}
\label{tab:oodshift}
\end{table}

\paragraph{What the audit cannot separate.}
The alarm tests for one region dominating the decision or for the instruction losing control of it; it does not test for a backdoor.
Three cases follow.
First, an alarm says that the implanted backdoor is firing, not which family placed the trigger on screen: the Grounding victim alarms on 0.896 of the steps carrying the TypeText overlay, whose icon it clicks on 0.899 of them (\Cref{tab:specificity}).
Second, an atypical clean step is flagged with probability 0.125--0.165 once it passes the gate, so a deployment with many atypical screens concentrates its false alarms on them.
Third, the false-positive guarantee needs test steps that are exchangeable with $B_2$.
When calibration and test share an unusual view, the rate stays near the target (the unresized collection of \Cref{tab:e04first}: 8 of 800 held-out clean steps at $\alpha=0.01$); when the test distribution moves and the calibration does not, no bound holds: the rate stayed at or below the target under the six shifts of \Cref{tab:oodshift}, which is an observation on those shifts and not a guarantee.
\section{Extra Baseline Detectors}
\label{sec:supp:extra}

\Cref{tab:extra_baselines} reports STRIP, SCALE-UP, an instruction-contrast probe, and a VLM judge under a separate per-victim protocol.
\Cref{tab:detectors} describes these four adaptations.
For each victim, we set the split-conformal threshold at a 1\% target FPR on 1000 clean calibration steps, and we evaluate the detector on 1000 additional clean steps and on the 1000 triggered steps of the victim's family.
Unlike \Cref{tab:strict}, the TPR here counts all triggered steps rather than only the followed ones, and the evaluation uses one fixed split instead of 20 app-disjoint seeds, so the two tables cannot be compared directly.
The agreement scores of SCALE-UP and of the instruction-contrast probe are discrete.
On every victim, their calibration threshold equals the maximum score, so neither raises an alarm on any clean or triggered step.
STRIP reaches 0.937 on Patch, but at a clean FPR of 0.011, above the target, and at most 0.169 elsewhere; the VLM judge reaches 0.481 on Popup and at most 0.028 elsewhere.

\begin{table*}[t]
\centering
\footnotesize
\setlength{\tabcolsep}{5pt}
\begin{tabular}{lccccc}
\toprule
& \multicolumn{5}{c}{TPR@1\% FPR / measured clean FPR} \\
\cmidrule(lr){2-6}
Detector & Patch & Popup & TypeText & Grounding & Adaptive-Blend \\
\midrule
STRIP~\cite{strip} & \textbf{0.937/0.011} & 0.031/0.002 & 0.000/0.005 & \textbf{0.169/0.004} & \textbf{0.058/0.009} \\
SCALE-UP~\cite{scaleup} & 0.000/0.000 & 0.000/0.000 & 0.000/0.000 & 0.000/0.000 & 0.000/0.000 \\
Instruction contrast & 0.000/0.000 & 0.000/0.000 & 0.000/0.000 & 0.000/0.000 & 0.000/0.000 \\
VLM judge & 0.000/0.003 & \textbf{0.481/0.005} & \textbf{0.028/0.004} & 0.003/0.002 & 0.000/0.002 \\
\midrule
AUROC (best TPR method) & 0.988 & 0.976 & 0.647 & 0.868 & 0.686 \\
\bottomrule
\end{tabular}
\caption{\textbf{Per-victim protocol: extra baseline detectors.} TPR@1\% FPR / measured clean FPR on one fixed split, counting all triggered steps; bold: best TPR per column; last row: AUROC of that detector.}
\label{tab:extra_baselines}
\vspace{-2mm}
\end{table*}

\section{Redirect Availability}
\label{sec:supp:redirect}

\begin{figure}[!htb]
\centering
\includegraphics[width=\columnwidth]{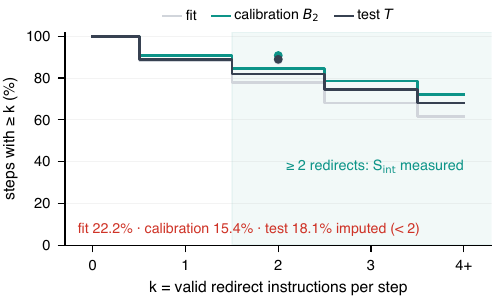}
\caption{\textbf{Valid redirects per step on the clean auxiliary set.} Fraction of steps with at least $k$ valid redirects under \cref{eq:capture}, per split; the band marks $k<2$, where $S_{\mathrm{int}}$ falls back to the clean median (red: its share).}
\label{fig:redirect}
\end{figure}

\Cref{fig:redirect} shows the distribution of valid redirects per step on clean AndroidControl steps.
Steps providing fewer than two valid redirects constitute a stratum where ICP defaults to the median intent scale $S_{\mathrm{int}}$ observed across all valid steps in $F$, as described in \cref{sec:icp}.

The joint rule treats these steps identically to any other clean step: they occupy the calibration split $B_2$ and are subject to
 the same gated threshold $T_\alpha$, ensuring the false-positive guarantee holds marginally across the entire clean distribution.
\section{Limitations}
\label{sec:supp:limit}

\paragraph{White-box access.}
IDEA reads the decision state of the checkpoint and ablates its visual tokens after the vision tower, so it requires the model weights.
A black-box deployment would need to approximate ICP and RCA through query access, paying a penalty in latency and statistic noise.

\paragraph{Diffuse triggers.}
RCA localizes compact regions at the resolution of merged visual tokens.
A diffuse, global trigger, \eg, a full-screen blended overlay or a whole-screen style shift, is not localized by element-level ablation because no single proposal removes it.
On Adaptive-Blend, the returned region is therefore not the trigger (mean IoU 0.016, \cref{sec:loc}), and detection rests on the excess-evidence side alone: only 19\% of its followed steps show an intent deficit.
IDEA reaches 0.898 TPR there on the episode-disjoint split at $\alpha=0.01$, on par with the Mahalanobis screen at 0.891 (\Cref{tab:strict}), and falls below the screen at $\alpha=0.005$ (0.569 vs.\ 0.713, \Cref{tab:strict_cp95}).
On the overlapping split, it reaches 0.983 against 0.949 at a higher clean FPR (\Cref{tab:main}).
We claim no advantage over the screen on this family.

\paragraph{Adaptive attackers.}
Our Adaptive-Blend family targets the trigger-as-region assumption, not the detection statistic.
It violates (A1) and weakens (A2) of \cref{prop:detect}: redirects override the blend, yet ablating one region still moves the decision state more than on clean steps, so it is caught by its excess evidence.
Because \cref{eq:rho} fires on either side, the audit misses a trigger that violates both assumptions, one that yields to redirects as clean steps do while no proposal reduces its effect.
An attacker who knows $\rho_{\mathrm{pc}}$ would therefore have to keep both the redirect effect and the largest region effect on the clean law at the triggered step's presence, \ie, $u\le 0$ and $v\le 0$, while still executing the target action.
We have not evaluated such an attacker and do not claim robustness against it.

\paragraph{Calibration domain.}
Our false-positive guarantee is marginal over clean steps exchangeable with the calibration split $B$.
It does not hold under shifts in app families or screen distributions.
Within the episode-disjoint split, atypical clean steps and steps from apps outside the victim's training pool are examined in \cref{sec:supp:ood}, together with benign shifts of the whole input distribution.
Outside that evidence, we make no claim about the false-positive rate under distribution shift.

\paragraph{Backbone coverage.}
All five-family evaluations use Qwen2.5-VL-3B-Instruct (\Cref{tab:strict}).
As shown in \Cref{tab:backbone}, Patch-only checks on Qwen2.5-VL-7B (0.999 TPR at 0.020 held-out clean FPR), UI-TARS-1.5-7B (0.921 at 0.023), and OS-Atlas-Base-7B (1.000 at 0.012) use the overlapping-split protocol.
The clean FPR is above the 1\% target on all three, and on UI-TARS IDEA is below the screen (1.000).
Because all three larger models derive from Qwen-VL bases (Qwen2.5-VL, or Qwen2-VL for OS-Atlas), transfer to other model families and to the remaining four attack families on larger backbones remains untested.
\section{Negative results and history of the statistic}
\label{sec:supp:negative}

This section records every statistic we screened and how the reported one was chosen. The operating points of \Cref{tab:strict} are thus not the result of an undisclosed search.
The candidates below, each with its acceptance screen, were fixed in writing before their runs and screened against the earlier statistic, the gated unconditional log-ratio $\rho=\log(E_{\max}+c)-\log(S_{\mathrm{int}}+c)$, which is now the ``$-$ conditioning'' row of \Cref{tab:ablation}. All of them failed their screens.

\paragraph{Q90-capped intent normalization.}
Replacing the clean median of $S_{\mathrm{int}}$ with a $q_{90}$ cap gains 0.448 TPR on Adaptive-Blend (0.513$\to$0.962) but costs 0.060 on Grounding (0.950$\to$0.890, bounding-box view 0.949$\to$0.889).
These deltas are measured against the log-ratio in the same fit configuration that IDEA uses (whitening and $S_{\mathrm{int}}$ imputation fit on $F$ only, literal $-\infty$ below the gate), on the 20 seeds of \Cref{tab:strict}. An earlier episode-disjoint run fitted the whitening and the $S_{\mathrm{int}}$ fallback median on all steps, including held-out test steps, and used a data-dependent sentinel; it gave 0.528 on Adaptive-Blend and 0.951 on Grounding. We disclose it as a corrected leak and report only the corrected configuration.
The screen required $\le$0.02 absolute loss on every other family, so this variant is rejected.

\paragraph{Whole-image Gaussian blur ($r{=}3$).}
Adding a fifth branch that blurs the entire screenshot before the vision tower is meant to catch non-local triggers.
It fails its clean-validity gate: clean task accuracy drops by 11.0, 10.0, 2.0, 6.0 and 6.5 percentage points across the five families, against a tolerance of 2 points fixed in advance, so the variant would trade detection for a victim it no longer measures.

\paragraph{Two-band DCT spectral erasure.}
Erasing two DCT bands before the vision tower was proposed as a non-local defense against blended triggers.
It fails the same clean-validity gate in its preliminary check: clean task accuracy drops by 2.0 and 2.5 on Patch, 5.0 and 3.5 on Popup, 1.0 and $-$3.0 on Grounding, 6.0 and 3.0 on Adaptive-Blend, and 0.5 and 1.0 points on TypeText at the low and high view, against a 2-point tolerance per view, so the check is recorded as a fail and no detection claim is made for it.
Because this variant fails its preliminary gate, it is not carried to the blind held-out test.

\paragraph{Max-fusion of the log-ratio with capped intent normalization.}
After the capped intent variant, we fixed in advance the maximum of the log-ratio and the capped ratio, each standardized by its median and median absolute deviation on $F$, under the same gate and one conformal threshold. This variant was designed after seeing the capped variant's development results.
The screen required at least 0.85 on Adaptive-Blend, at most 0.02 absolute loss on every other family and a mean clean FPR of at most 0.015.
This fusion reaches 0.962 on Adaptive-Blend at a mean pooled FPR$_T$ of 0.007 but inherits the capped variant's loss on Grounding (0.950$\to$0.891, bounding-box view 0.949$\to$0.890), so it is rejected.

\paragraph{Fusion of the log-ratio with the screen.}
Three rules that fuse the gated log-ratio $S$ with a second score were fixed in writing before their runs, together with the rule that none replaces the primary statistic.
The first is a Bonferroni union: alarm if $s_0$ exceeds its conformal threshold on $B_1\cup B_2$ at $\alpha/2$ or $S$ exceeds its threshold on $B_2$ at $\alpha/2$.
The second is a slot fusion: each score is centered and scaled by its $\alpha/2$ threshold, the larger of the two is taken, and one threshold at $\alpha$ is set on $B_2$.
At $\alpha=0.01$ over the 20 episode-disjoint seeds, the Bonferroni union and slot fusion reach 0.834 and 0.887 on Adaptive-Blend (seed sd 0.214 and 0.180, worst seed 0.253 for both) and 0.801 and 0.834 on Grounding (bounding-box view 0.798 and 0.832), against 0.513 and 0.950 for the log-ratio; both stay at 1.000 on Patch, Popup and TypeText.
Pooled FPR$_T$ is 0.006--0.008 for the Bonferroni union and 0.008--0.011 for the slot fusion.
Neither has an operating point at $\alpha=0.005$: the $\alpha/2$ quantile needs at least 399 calibration steps and $B_2$ holds 246--359.
The third rule is the Bonferroni union with $E_{\max}$ in place of the screen. On the first episode-disjoint run, before the fit was restricted to $F$ (a correction that moved the log-ratio by at most 0.015), it reached 0.880 on Adaptive-Blend and 0.774 on Grounding, below the slot fusion on both, and it was not rerun.
All three are below IDEA on Grounding (0.962) and on Adaptive-Blend (0.898) and are not used.

\paragraph{History of the statistic.}
The method of this paper was first fixed with the gated log-ratio as its statistic, together with a clause that no further variant would be tried after the max-fusion candidate.
After the results above we set that clause aside by our own decision and replaced the log-ratio with the presence-conditional statistic of \cref{eq:rho}.
This choice is post hoc with respect to the five families of \Cref{tab:strict}: the rule was designed after inspecting their triggered steps, so that table is development evidence for it, not a blind test.
In the same pass we examined $\max(u,v)$, $u+v$, $u$ alone, $v$ alone and the ungated statistic, which are the rows of \Cref{tab:ablation}, and four variants that we did not carry forward: regressing on $s_0$ instead of $n_0$, setting the intent deficit to zero when $S_{\mathrm{int}}$ is not measured, a three-dimensional Mahalanobis distance on $(\log E_{\max},\log S_{\mathrm{int}},\log n_0)$, and two ratio forms (radial shares and an elasticity exponent).
The quadrature form was preferred to $\max(u,v)$ on the $\alpha=0.005$ column (Patch 0.978 vs.\ 0.904); the two agree within 0.003 at $\alpha=0.01$.
Before any triggered step of the held-out victims of \cref{sec:blind} was scored, we froze \cref{eq:rho} with a code hash and fixed three conditions: no family more than 0.02 below the log-ratio at $\alpha=0.01$ with mean FPR$_T$ at most 0.015 (met, \Cref{tab:strict_cp95}); the held-out victims reported for IDEA, the log-ratio, the screen and $E_{\max}$ on the same steps whatever the outcome; and every earlier table recomputed and reported as measured, including where IDEA loses to the log-ratio (the $\alpha=0.005$ column of \Cref{tab:strict_cp95} and UI-TARS in \Cref{tab:backbone}).
No variant was tried after the freeze.
\section{Held-out blend: first collection and its correction}
\label{sec:supp:e04first}

The trigger, the cohort and the attack-quality criteria of the held-out blend victim of \cref{sec:blind} were fixed before training.
Its detector inputs were first collected by a launch that omitted the screenshot resizing (a cap of 1{,}003{,}520 pixels) applied in the victim's training, in its attack-quality evaluation and in every other collection of this paper.
At the unresized view the victim follows none of the 1000 triggered steps, and its clean step success in the collection is 0.601, against a clean accuracy of 0.896 in the attack-quality evaluation.
The set of followed steps is therefore empty, and TPR, AUROC and the paired bootstrap are undefined; we make no detection claim from this run.
It does measure false alarms when calibration and test data share a view that the victim was not trained on (\Cref{tab:e04first}).
All four scores stay near the target on held-out clean steps.
On the ignored triggered steps IDEA alarms more often than the log-ratio, the screen and $E_{\max}$ (0.073 against 0.053, 0.044 and 0.053 at $\alpha=0.01$), and 0.129 of these steps pass the gate, against 0.043 of the clean test steps: the overlay moves the decision state although the backdoor does not fire.

\begin{table}[!htb]
\centering
\footnotesize
\setlength{\tabcolsep}{3pt}
\begin{tabular}{@{}llccc@{}}
\toprule
Score & $\alpha$ & Ignored trig. & FPR$_T$ (CP95) & FPR, clean test \\
\midrule
IDEA & 0.01 & 73/1000 & 8/800 [0.004, 0.020] & 8/1000 \\
Log-ratio & 0.01 & 53/1000 & 6/800 [0.003, 0.016] & 5/1000 \\
Screen & 0.01 & 44/1000 & 10/800 [0.006, 0.023] & 9/1000 \\
$E_{\max}$ & 0.01 & 53/1000 & 12/800 [0.008, 0.026] & 7/1000 \\
\midrule
IDEA & 0.005 & 56/1000 & 4/800 [0.001, 0.013] & 1/1000 \\
Log-ratio & 0.005 & 43/1000 & 4/800 [0.001, 0.013] & 5/1000 \\
Screen & 0.005 & 19/1000 & 3/800 [0.001, 0.011] & 5/1000 \\
$E_{\max}$ & 0.005 & 34/1000 & 6/800 [0.003, 0.016] & 6/1000 \\
\bottomrule
\end{tabular}
\caption{\textbf{Held-out blend, first collection (no resizing): alarm counts.} No triggered step is followed, so no TPR is defined; FPR$_T$ on the held-out clean split [CP95] and on the 1000 clean test steps.}
\label{tab:e04first}
\end{table}

Before repeating the collection we fixed the following in writing: the same command with only the resizing restored, written to a new directory; the same victim, trigger set, cohort, seed and frozen code (hashes checked before and after); scores opened only if the repeated collection itself reaches an ASR of at least 0.80 with at least 500 followed steps, and the cell reported as inconclusive otherwise; the outcome reported next to the first collection whatever it is; and no further collection, trigger, level or seed tried afterwards.
The repeated collection is the one reported in \cref{sec:blind}.

\section{Component ablation at the tighter target}
\label{sec:supp:ablation}

\Cref{tab:ablation} in the main text removes or replaces one component of the gated statistic at a time at $\alpha=0.01$ and gives the five-family average at $\alpha=0.005$; \Cref{tab:ablation005} gives the per-family values at $\alpha=0.005$.
Every row keeps the same forwards, fit and split-conformal calibration.
At this level the threshold rests on one or two calibration scores (\cref{sec:calib}), and the order of the rules changes.
The log-ratio and $u+v$ are ahead of IDEA on Patch, Popup, TypeText and Grounding (0.761 and 0.810 vs.\ 0.702 on Grounding), and $v$ alone is ahead on Grounding (0.774).
IDEA keeps the highest average (0.844) because these rules lose more on Adaptive-Blend, where the screen (0.713) is ahead of every gated rule except $u$ alone (0.760).

\begin{table}[!htb]
\centering
\footnotesize
\setlength{\tabcolsep}{2.6pt}
\resizebox{\columnwidth}{!}{%
\begin{tabular}{lcccccc}
\toprule
Variant & Patch & Popup & Ground. & Adapt. & TypeText & AVG \\
\midrule
IDEA (gated $\rho_{\mathrm{pc}}$) & 0.978 & 0.996 & 0.702 & 0.569 & 0.976 & \best{0.844} \\
\midrule
$-$ gate (ungated $\rho_{\mathrm{pc}}$) & 0.978 & 0.996 & 0.651 & 0.491 & 0.974 & 0.818 \\
$-$ conditioning (gated log-ratio) & \best{1.000} & \best{1.000} & 0.761 & 0.313 & \best{0.997} & 0.814 \\
$u$ only (gated) & 0.900 & 0.952 & 0.170 & \best{0.760} & 0.799 & 0.716 \\
$v$ only (gated) & 0.882 & 0.959 & 0.774 & 0.137 & 0.907 & 0.732 \\
$u+v$ (gated) & \best{1.000} & \best{1.000} & \best{0.810} & 0.253 & \best{0.997} & 0.812 \\
$\max(u,v)$ (gated) & 0.904 & 0.959 & 0.673 & 0.562 & 0.919 & 0.803 \\
\midrule
gate only (Mahalanobis screen) & 0.983 & 0.977 & 0.276 & 0.713 & 0.991 & 0.788 \\
\bottomrule
\end{tabular}%
}
\caption{\textbf{Component ablation at $\alpha=0.005$ on the strict split.} TPR on followed steps, mean over 20 seeds, and the five-family average. Bold: best per column. Ground.: Grounding; Adapt.: Adaptive-Blend.}
\label{tab:ablation005}
\end{table}

\section{Proposal count}
\label{sec:supp:topm}

\Cref{tab:topm} reruns the frozen rule with the candidate set cut to the top-$M$ attribution-ranked proposals, $M\in\{1,2,3,4\}$, on the stored hidden states of the episode-disjoint split (20 seeds, 4000 rows per family).
Forwards, whitening, clean law and both thresholds are unchanged; only the set over which $\max_j E_j$ is taken changes.
At $M{=}4$ the table reproduces \Cref{tab:strict} to rounding, which confirms the replay against the collected results.
Reducing $M$ is close to free on four families: Patch, Popup, TypeText and Grounding move by at most 0.003 in TPR and by at most 0.001 in pooled FPR$_T$, and on Grounding the $R^\star$ hit rate even rises from 0.910 at $M{=}4$ to 0.945 at $M{=}1$, because the single top-ranked proposal is the trigger more often than the argmax over four.
Adaptive-Blend is the exception: its TPR rises monotonically with $M$, from 0.827 at $M{=}1$ to 0.898 at $M{=}4$, because the blend has no single trigger region and the evidence is spread across proposals; this is the family the extra proposals are kept for.

\begin{table}[!htb]
\centering
\footnotesize
\setlength{\tabcolsep}{5pt}
\renewcommand{\arraystretch}{0.95}
\begin{tabular}{@{}lcccccc@{}}
\toprule
$M$ & Patch & Popup & Ground. & Adapt. & TypeText & AVG \\
\midrule
1 & 1.000 & 1.000 & 0.965 & 0.827 & 1.000 & 0.958 \\
2 & 1.000 & 1.000 & 0.963 & 0.886 & 0.999 & 0.970 \\
3 & 1.000 & 1.000 & 0.963 & 0.894 & 0.999 & 0.971 \\
4 & 1.000 & 1.000 & 0.962 & 0.898 & 0.999 & \best{0.972} \\
\bottomrule
\end{tabular}
\caption{\textbf{Proposal count.} TPR on followed steps at $\alpha=0.01$ with the candidate set cut to the top-$M$ ranked proposals, replay on stored hidden states of the strict split, mean over 20 seeds. AVG over the five families.}
\label{tab:topm}
\end{table}

\section{Candidate sets: attribution proposals versus token grids}
\label{sec:supp:grid}

\noindent\textbf{What this tests.}
A reviewer can ask why the audit scores four attribution-ranked proposals instead of every region, as a detector that tiles the input would.
This section replaces the candidate set that defines $E_{\max}=\max_j E_j$ and keeps everything else frozen: the fit on $F$, the gate threshold $T_1$, the clean law and its two coefficients, and the split-conformal threshold $T_\alpha$ on $B_2$.
Only the set of regions whose visual tokens are replaced by the blur tokens changes.

\noindent\textbf{Sets.}
\emph{top-4} is the configuration of \cref{sec:rca}: the four highest-ranked attribution proposals.
\emph{$4{\times}4$} and \emph{$8{\times}8$} tile the screenshot into 16 and 64 cells and ignore attribution entirely.
The two mixed sets append the grid cells to the attribution proposals, so $E_{\max}$ is taken over $4{+}16$ or $4{+}64$ regions.

\noindent\textbf{Result.}
Patch and Popup detect 0.999--1.000 under every set, so the choice does not matter there.
On Grounding the grid-only sets lose 0.06--0.07 (0.906 for $4{\times}4$, 0.893 for $8{\times}8$, against 0.962) and never localize ($R^\star$ hit at most 0.388, IoU$>0.5$ = 0.000), while the attribution proposals keep hit 0.910.
On Adaptive-Blend the $4{\times}4$ grid is better than the attribution set (0.969 against 0.898), as expected for a trigger that covers the screen.
Pooled FPR$_T$ stays at 0.005--0.009 for every set.
We therefore keep four attribution-ranked proposals: they are the only set that both detects every family and localizes the trigger.

\begin{table}[!htb]
\centering
\scriptsize
\setlength{\tabcolsep}{4pt}
\resizebox{\columnwidth}{!}{%
\begin{tabular}{@{}lccccc@{}} \toprule
Candidate set & Patch & Popup & Ground. & Adapt. & TypeText \\ \midrule
top-4 attribution ($M{=}4$) & 1.000 & \best{1.000} & 0.962 & 0.898 & 0.999 \\
$4{\times}4$ grid only & 0.999 & \best{1.000} & 0.906 & 0.969 & 0.999 \\
$8{\times}8$ grid only & 0.999 & \best{1.000} & 0.893 & 0.964 & 0.999 \\
top-4 + $4{\times}4$ & 0.999 & \best{1.000} & 0.931 & 0.956 & 0.998 \\
top-4 + $8{\times}8$ & 0.999 & \best{1.000} & 0.950 & 0.915 & 0.999 \\
\midrule
\multicolumn{1}{@{}l}{\emph{seed sd}} & 0.004 & 0.000 & 0.082 & 0.091 & 0.004 \\
\multicolumn{1}{@{}l}{\emph{pooled FPR$_T$}} & 0.005--0.006 & 0.008--0.009 & 0.006--0.008 & 0.005--0.007 & 0.006--0.006 \\
\bottomrule \end{tabular}%
}

\caption{\textbf{Candidate sets under the frozen rule.} TPR on followed steps at $\alpha=0.01$, mean over 20 seeds, with the seed standard deviation and the pooled FPR$_T$ range over the five sets of each family. Ground.: Grounding; Adapt.: Adaptive-Blend.}
\label{tab:grid}
\vspace{-2mm}
\end{table}

\section{Victim Checkpoint Quality}
\label{sec:supp:victim}

\Cref{tab:victim} reports the quality of the five backdoored checkpoints, each evaluated on 1000 clean and 1000 triggered held-out steps with greedy decoding.
All five keep the clean step success within 2 points of the clean reference and reach an attack success rate (ASR) of at least 0.98.
The ASR column comes from this victim-quality evaluation, whose success rule accepts a click inside the trigger box for Popup and Grounding; the detection experiments instead label followed steps from the predictions of the audited run itself, which uses a different software stack and a stricter rule (exact target string for TypeText, click within 0.02 of the target point for Grounding), and this gives the 858 and 908 followed steps of \cref{sec:exp} (997 for Grounding under the in-box rule).

\paragraph{Training.}
Every victim is Qwen2.5-VL-3B-Instruct fine-tuned with LoRA (rank 16, scaling 32, dropout 0.05) on all linear layers of the language model and the vision tower, with the token embeddings, the output head and the visual merger trained in full.
Training runs for 2 epochs at learning rate $10^{-4}$ with a cosine schedule, warmup ratio 0.1 and bf16 precision; screenshots are resized to at most 1{,}003{,}520 pixels and sequences are cut at 4096 tokens.
Each training set is built from 13{,}603 steps, of which 1{,}360 (10\%) carry the trigger and the attacker's target action.
The trigger is 128\,px for Patch, TypeText and Grounding; the Adaptive-Blend overlay has opacity 0.15, and its redirect augmentation adds records that pair a triggered screenshot with a redirected instruction and the clean ground-truth action.

\begin{table}[!htb]
\centering
\footnotesize
\setlength{\tabcolsep}{4pt}
\begin{tabular}{lcc}
\toprule
Family & Clean step success & ASR \\
\midrule
Clean reference & 0.858 & 0.071 \\
\midrule
Patch & 0.868 & 0.999 \\
Popup & 0.867 & 1.000 \\
TypeText & 0.864 & 0.988 \\
Grounding & 0.863 & 0.994 \\
Adaptive-Blend & 0.856 & 0.981 \\
\bottomrule
\end{tabular}
\caption{\textbf{Victim checkpoint quality.} Clean step success and ASR on 1000 held-out steps each; all five meet the pre-specified criteria (drop $\le 2$\,pt, ASR $\ge 0.80$). Clean reference: chance rate.}
\label{tab:victim}
\vspace{-2mm}
\end{table}

\section{Backbone sensitivity}
\label{sec:supp:backbone}

\Cref{tab:backbone} repeats the Patch experiment on three 7B backbones under the single-split protocol of \cref{sec:supp:overlap}, the only protocol measured on them; on the strict split the 3B Patch values are 1.000 for IDEA and 0.999 for the screen (\Cref{tab:strict}).

\begin{table*}[t]
\centering
\footnotesize
\setlength{\tabcolsep}{2.5pt}
\begin{tabular}{lcccccccc}
\toprule
& & & & \multicolumn{5}{c}{TPR@1\% FPR on followed steps / held-out clean FPR} \\
\cmidrule(lr){5-9}
Backbone & $n_{\mathrm{fit}}/d$ & ASR & Clean acc. & IDEA (gated $\rho_{\mathrm{pc}}$) & \begin{tabular}[b]{@{}c@{}}Log-ratio\\(ablation)\end{tabular} & \begin{tabular}[b]{@{}c@{}}Earlier IDEA score\\(superseded)\end{tabular} & Mahalanobis screen & Best other baseline \\
\midrule
Qwen2.5-VL-3B & 998/2048 & 0.999 & 0.868 & 1.000/0.017 & 1.000/0.011 & 0.015/0.006 & 1.000/0.011 & OC-SVM 1.000/0.014 \\
Qwen2.5-VL-7B & 998/3584 & 1.000 & 0.858 & 0.999/0.020 & 0.999/0.013 & 0.109/0.005 & 1.000/0.014 & OC-SVM 0.994/0.009 \\
UI-TARS-1.5-7B & 998/3584 & 1.000 & 0.890 & 0.921/0.023 & 0.996/0.026 & 0.221/0.006 & 1.000/0.013 & OC-SVM 0.999/0.012 \\
OS-Atlas-Base-7B & 998/3584 & 0.998 & 0.812 & 1.000/0.012 & 1.000/0.013 & 0.078/0.014 & 0.999/0.009 & OC-SVM 0.991/0.010 \\
\bottomrule
\end{tabular}
\caption{\textbf{Backbone sensitivity on Patch} (overlapping single split, $n_{\mathrm{fit}}=998$, 1\% target). IDEA: frozen rule; ``earlier IDEA score'': superseded whitened-space variant. OS-Atlas is Qwen2-VL-based, the others Qwen2.5-VL.}
\label{tab:backbone}
\vspace{-2mm}
\end{table*}

As shown in \Cref{tab:backbone}, replayed on the stored Qwen2.5-VL-7B states, our method detects 996 of the 997 followed Patch steps (TPR 0.999) at a held-out clean FPR of 0.020, compared with 0.999 TPR at 0.013 FPR for the log-ratio and 1.000 TPR at 0.014 FPR for the Mahalanobis screen.
The 0.109 TPR of an earlier whitened-space variant (0.015 on 3B) reflects that superseded score rather than the backbone.
On UI-TARS-1.5-7B, IDEA detects 921 of 1000 steps at a clean FPR of 0.023, falling below the log-ratio (0.996 TPR at 0.026 FPR), the Mahalanobis screen (1.000 TPR at 0.013 FPR), and the one-class SVM (0.999 TPR at 0.012 FPR).
This is because the clean law fit on $B_1$ transfers less well to this backbone, where 9.6\% of clean test steps pass the gate against 5.2--5.3\% on the other two.
On OS-Atlas-Base-7B, a Qwen2-VL derivative, IDEA detects all 999 followed steps at a clean FPR of 0.012, compared with 0.999 TPR at 0.009 FPR for the Mahalanobis screen.
On every backbone, including 3B (0.017), the held-out clean FPR of IDEA exceeds the 1\% target although its FPR on the calibration half $B_2$ is 0.008.
This is because the overlapping-split calibration does not transfer exactly to the test steps, so no row of \Cref{tab:backbone} is a comparison at equal FPR.
Nothing in the method was retuned for any 7B backbone.
\section{Cost}
\label{sec:supp:cost}

\Cref{tab:cost} times the frozen rule online, step by step, on the Patch and Grounding victims (Qwen2.5-VL-3B, bf16, batch 1, one GPU).
We timed 300 clean and 302 (Patch) or 392 (Grounding) triggered test steps twice, once with the early exit of the gate and once with the audit forced on every step, with CUDA synchronized at every stage boundary and warm-up excluded.
The online scores reproduce the stored offline ones, as gate and alarm decisions agree on all 602 and 692 timed steps at both $\alpha$, confirming that the timed rule is the evaluated one.
The agent step itself (screenshot loading, vision encoding, action generation) takes 0.65\,s.
The gate adds one decision-state pass of a single row over the already encoded screenshot plus the Mahalanobis distance, taking 0.034\,s.
Steps that pass the gate also run the audit, taking 0.79\,s in median.
This audit comprises proposal ranking with attribution (0.08\,s), the $K$ teacher-forced validity checks of \cref{eq:capture} (0.21\,s), one vision-tower encoding of the blurred screenshot (0.30\,s), and one batched decision-state pass over the $K{+}M$ redirected and ablated inputs with the score (0.20\,s).
With the early exit, 2.7\% (Patch) and 3.0\% (Grounding) of clean steps pass the gate, and a clean step takes 0.672 and 0.677\,s (p95 0.720 and 0.730\,s), a mean overhead of 0.054 and 0.057\,s, or 9.3\% and 9.8\% of the agent step (\Cref{tab:cost}).
Forcing the audit on every step raises this time to 1.40\,s (p95 1.50--1.51\,s).
Triggered steps mostly pass the gate (100\% on Patch, 90.3\% on Grounding) and take 1.41--1.44\,s, so a step that IDEA flags costs about two agent steps, and the check always finishes before the action is executed.
TeCo re-runs the agent under up to 25 corrupted screenshots (five corruption types at five severities) and takes 15.4 and 12.9\,s per clean step, which is 19--23 times the clean step under IDEA.
The Mahalanobis screen timed alone, with its own encoding, takes 0.33--0.35\,s.
Run beside the agent, it reuses the encoding and costs what the gate costs.
The remaining baselines were timed earlier inside the evaluation pipeline on cached encodings and over five families, so their rows give detector-only times that exclude the agent step (\Cref{tab:cost}).
Specifically, IBD-PSC runs five forwards with scaled normalization weights, DeDe runs one decoder pass after the encoder, Entropy runs two paraphrased generations plus the scoring of the candidate actions, and Margin reuses those candidates.
The Grounding run had its GPU to itself with no other process before or after.
The Patch force and early-exit passes also ran alone, but another user's inference server started on the same GPU during the Patch TeCo pass, which can only inflate that TeCo time.

\begin{table}[!htb]
\centering
\footnotesize
\setlength{\tabcolsep}{4pt}
\begin{tabular}{lcc}
\toprule
 & Patch & Ground. \\
\midrule
\multicolumn{3}{l}{\emph{Whole step, timed online}} \\
Agent alone & 0.647 & 0.646 \\
IDEA, clean step, early exit & 0.672 & 0.677 \\
IDEA, clean step, audit forced & 1.403 & 1.398 \\
IDEA, triggered step, early exit & 1.414 & 1.435 \\
\midrule
\multicolumn{3}{l}{\emph{Detector only, timed online}} \\
IDEA gate & 0.034 & 0.034 \\
IDEA audit (gate passed) & 0.787 & 0.785 \\
Mahalanobis, own encoding & 0.333 & 0.348 \\
TeCo & 15.42 & 12.95 \\
\midrule
\multicolumn{3}{l}{\emph{Detector only, evaluation pipeline, range over five families}} \\
IBD-PSC & \multicolumn{2}{c}{0.222--0.265} \\
DeDe & \multicolumn{2}{c}{0.050--0.076} \\
Entropy & \multicolumn{2}{c}{1.170--1.711} \\
Margin & \multicolumn{2}{c}{0.176--0.268} \\
\bottomrule
\end{tabular}
\caption{\textbf{Cost per step} (median seconds, one GPU, Qwen2.5-VL-3B). Whole-step rows include the agent; detector-only rows do not. Protocol in \cref{sec:supp:cost}.}
\label{tab:cost}
\vspace{-2mm}
\end{table}

\section{Algorithm}
\label{sec:supp:alg}

This
 section restates \cref{alg:idea} in detail, specifying the mapping to our method equations.

\textbf{Intent-Dominance Evidence Audit (IDEA).}
Given an agent $\pi$, the decision-state readout $h(\cdot)$, the whitening $z(\cdot)$ estimated on split $F$, the clean law parameters fit on $B_1$, a test step $(x, I)$, and thresholds $T_1$ (conformal on $B_1$) and $T_\alpha$ (on $B_2$):
\begin{enumerate}[
leftmargin=1.2em,itemsep=1pt,topsep=2pt,parsep=0pt]
\item \textbf{Gate.} Compute $h_0 = h(x, I)$ and its
 Mahalanobis distance $s_0$ from the clean mean. If $s_0 \le T_1
$, assign $S = -\infty$ and skip to step 5.
\item \textbf{Intent.} Identify
 $K=4$ redirect instructions $I'_1, \ldots, I'_K$ naming visible elements other than
 the one named by $I$. Retain the $K'$ instructions that satisfy \cref{eq:capture}.
 Compute $S_{\mathrm{int}}$ as the median whitened displacement $\|z(h(x, I'_k)) - z(h_0)\|$ (\cref{eq:sint}). If $K'<2$,
 replace $S_{\mathrm{int}}$ with the median value calculated over the valid steps of $F$.
\item
 \textbf{Evidence.} Generate $M=4$ candidate regions $R_j$ from accessibility-tree boxes,
 OCR boxes, and gradient-based attribution, supplemented by a fixed grid. Replace the visual tokens of each $R_j$ with those of a 3\,px blurred screenshot $\tilde{x}$, re-encode $x_{-R
_j}$, and compute the whitened displacement $E_j = \|z(h_0) - z(h(x_{-R_j}, I))\|$ (\cref{eq:ej}).
\item \textbf{Dominance.} With $E_{\max} = \max_j E_j$, current presence $n_
0=\|z(h_0)\|$, and the precomputed floor $c$, extract residuals $u$ and $v$
 from the clean law regression on $B_1$ (\cref{eq:law}). Set the statistic $S
 = \rho_{\mathrm{pc}} = \sqrt{\max(u,0)^2+\max(v,0)^2}$ (\cref{eq:rho}).
\item \textbf{Threshold.} If $S >
 T_\alpha$ (\cref{eq:joint}), raise an alarm and return the hypothesis $R^\star =
 R_{\arg\max_j E_j}$ alongside the action $a$. Otherwise, classify the step as clean. Split-conformal calibration on $B_2$ ensures that $\Pr(\text{alarm} \mid \text{clean}) \le \alpha$.
\end{enumerate}

\section{Requirements Comparison}
\label{sec:supp:req}

\Cref{tab:req} compares published backdoor detectors with the requirements of our threat model, according to the scope that
 each paper defines.
In our experiments every baseline, whether adapted from its original form or not, uses a split-conformal threshold
 and thus respects the same marginal bound on its false-positive rate at the target level.
The final column of the table does not describe
 our adapted implementations: it records whether the original method, as published, specifies a calibrated finite-sample FPR bound.

\begin{table}[!htb]
\centering
\small
\setlength{\tabcolsep}{1.8pt}
\begin{tabular}{lcccccc}
\toprule
Method & Cal. & Step & Agnostic & Action & Loc. & FPR guar. \\
\midrule
STRIP & $\checkmark$ & $\checkmark$ & $\checkmark$ & $\times$ & $\times$ & $\times$ \\
TeCo & $\checkmark$ & $\checkmark$ & $\checkmark$ & $\times$ & $\times$ & $\times$ \\
SCALE-UP & $\checkmark$ & $\checkmark$ & $\checkmark$ & $\times$ & $\times$ & $\times$ \\
IBD-PSC & $\checkmark$ & $\checkmark$ & $\checkmark$ & $\times$ & $\times$ & $\times$ \\
DeDe & $\checkmark$ & $\checkmark$ & $\checkmark$ & $\times$ & $\times$ & $\times$ \\
Mahalanobis & $\checkmark$ & $\checkmark$ & $\times$ & $\checkmark$ & $\times$ & $\times$ \\
\textbf{IDEA} & $\checkmark$ & $\checkmark$ & $\checkmark$ & $\checkmark$ & $\checkmark$ & $\checkmark$ \\
\bottomrule
\end{tabular}
\caption{\textbf{Requirements of the threat model, by the stated scope of each method.} Cal.: clean-only calibration; Step: per-step decision; Agnostic: attack-agnostic; Action: GUI action output; Loc.: returns a region; FPR guar.: finite-sample FPR bound as published.}
\label{tab:req}
\vspace{-2mm}
\end{table}

\begin{table*}[t]
\centering
\footnotesize
\begin{tabular}{lp{0.74\linewidth}}
\toprule
Detector & Adaptation \\
\midrule
\multicolumn{2}{l}{\emph{Strict split (\Cref{tab:strict}); all but the last five also on the earlier single split (\Cref{tab:main})}} \\
TeCo~\cite{teco} & Re-runs the agent under five corruption types (Gaussian noise, Gaussian blur, JPEG, brightness, contrast) at five severities; score = standard deviation over types of the first severity that changes the action \\
IBD-PSC~\cite{ibdpsc} & Normalization weights of the last $k$ decoder layers scaled by 1.5 for $k=1,\dots,5$; score = mean probability of the original first action token \\
DeDe~\cite{dede} & A decoder is trained to reconstruct the screenshot from the visual tokens; score = pixel error against the $8\times$ downsampled screenshot \\
Mahalanobis screen~\cite{lee2018mahalanobis} & Mahalanobis distance of $h_0$ under the empirical covariance of $F$ (the gate of IDEA, used alone) \\
Entropy & Entropy of the model's distribution over candidate actions (the agent's action and its actions under two instruction paraphrases) \\
Margin & Top-1 minus top-2 probability over the same candidate actions \\
Random occlusion & Four random black patches (side 0.2 of each image dimension); score = largest drop in the log-probability of the action \\
Area occlusion & One random region per IDEA proposal with the same number of visual tokens, replaced by blurred tokens; score = largest drop in the log-probability of the action \\
LW-Mahalanobis & Norm of $h_0$ after whitening with the Ledoit--Wolf shrunk covariance of $F$ \\
$k$NN-10 & Mean distance of $h_0$ to its 10 nearest neighbors in $F$, in the whitened space \\
$z$-score max & Largest absolute whitened coordinate of $h_0$ \\
Cosine spectrum & Mean cosine similarity between $h_0$ and the decision states under the valid redirects \\
PCA residual & Norm of the part of $h_0$ outside the top 64 principal directions of $F$ \\
\midrule
\multicolumn{2}{l}{\emph{Earlier single split only (\Cref{tab:main,tab:backbone})}} \\
OC-SVM~\cite{schoelkopf2001ocsvm} & One-class SVM (RBF kernel, $\nu=0.1$) on $h_0$, fit on the fit split \\
SCALE-UP~\cite{scaleup} & Whole-image pixel multiplication at scales 1--5; score = share of scaled copies whose action differs from the original \\
\midrule
\multicolumn{2}{l}{\emph{Per-victim protocol (\Cref{tab:extra_baselines})}} \\
STRIP~\cite{strip} & Ten blends at weight 0.5 with clean calibration screenshots; score = negative mean entropy over the first tokens of the nine action types \\
SCALE-UP~\cite{scaleup} & Pixel values multiplied by 2, 3, 4 and 5; score = mean agreement with the original action (same action type, click within 50\,px or in the same target box, same typed text) \\
Instruction contrast & Four fixed rewordings of the instruction; score = mean agreement of the actions over the six pairs. Inspired by text-contrast detection for CLIP, not a reproduction of a published detector \\
VLM judge & The victim is asked whether the screenshot or the proposed action is suspicious; score = logit of ``yes'' minus logit of ``no''. A self-judge heuristic, not a published detector \\
\bottomrule
\end{tabular}
\caption{\textbf{Baseline detectors and their adaptations.} $h_0$: decision state of the step; $F$: fit split; LW: Ledoit--Wolf shrinkage.}
\label{tab:detectors}
\end{table*}

\begin{table*}[t]
\centering
\footnotesize
\setlength{\tabcolsep}{3pt}
\begin{tabular}{llcccccccccc}
\toprule
& & \multicolumn{4}{c}{Strict split, IDEA alarms, 20 seeds, mean} & \multicolumn{3}{c}{Overlapping, RCA (ours)} & \multicolumn{3}{c}{Overlapping, earlier variant} \\
\cmidrule(lr){3-6} \cmidrule(lr){7-9} \cmidrule(lr){10-12}
Family & Rule & Hit & IoU$>0.5$ & mIoU & H/I/m at $\alpha{=}0.005$ & Hit & IoU$>0.5$ & mIoU & Hit & IoU$>0.5$ & mIoU \\
\midrule
Patch & point & 0.862 & 0.673 & 0.546 & 0.863/0.673/0.547 & 0.859 & 0.670 & 0.544 & 0.295 & 0.205 & 0.193 \\
Popup & bbox & 0.993 & 0.000 & 0.160 & 0.993/0.000/0.160 & 0.994 & 0.000 & 0.160 & 0.536 & 0.000 & 0.082 \\
Grounding & point & 0.910 & 0.138 & 0.318 & 0.910/0.147/0.323 & 0.826 & 0.121 & 0.289 & 0.589 & 0.068 & 0.201 \\
Grounding$^\dagger$ & bbox & 0.914 & 0.137 & 0.320 & 0.909/0.146/0.323 & \multicolumn{3}{c}{not run} & \multicolumn{3}{c}{not run} \\
Adaptive-Blend & point & 1.000 & 0.011 & 0.016 & 1.000/0.016/0.021 & 1.000 & 0.013 & 0.017 & \multicolumn{3}{c}{not run} \\
TypeText & point & 0.494 & 0.323 & 0.428 & 0.494/0.323/0.429 & 0.495 & 0.325 & 0.429 & 0.369 & 0.228 & 0.329 \\
\bottomrule
\end{tabular}
\caption{\textbf{Localization.} Strict split: correct alarms of the frozen rule at $\alpha=0.01$ (and $0.005$), mean over 20 seeds; overlapping split: one-seed earlier run. Hit: center of $R^\star$ inside the trigger box. $^\dagger$Scored on the log-ratio's alarms.}
\label{tab:loc}
\end{table*}

\section{Tail Metrics}
\label{sec:supp:tail}

\Cref{tab:tail} reports TPR at stricter operating points (0.5\% and 0.1\% target FPR) for two ungated scores of the earlier episode-disjoint pipeline: the Mahalanobis screen and the unconditional log-ratio $\rho$ (the ablation statistic), at its calibration size ($|F|{=}998$, $|
B|{=}1002$).
Both scores were swept at both levels for Grounding, Adaptive-Blend
 and TypeText only, so the table displays these three families.
The gated rule of \cref{eq:joint} is measured at $\alpha=0.01$ and $0.005$
 on the strict split (\Cref{tab:strict_cp95} in the supplement) and is not swept
 at 0.1\%, a rate its calibration split cannot certify.
At 0.1\% the Mahalanobis screen retains a tail
 on Adaptive-Blend (0.744) where the log-ratio collapses (0.026),
 and the log-ratio retains one on Grounding (0.256) where the screen weakens (0.090).

\begin{table}[!htb]
\centering
\footnotesize
\setlength{\tabcolsep}{3pt}
\begin{tabular}{lcccccc}
\toprule
& \multicolumn{2}{c}{Grounding} & \multicolumn{2}{c}{Adaptive-Blend} & \multicolumn{2}{c}{TypeText} \\
\cmidrule(lr){2-3} \cmidrule(lr){4-5} \cmidrule(lr){6-7}
& @0.5\% & @0.1\% & @0.5\% & @0.1\% & @0.5\% & @0.1\% \\
\midrule
Mahalanobis & 0.630 & 0.090 & 0.903 & 0.744 & 1.000 & 1.000 \\
Log-ratio $\rho$ & 0.456 & 0.256 & 0.063 & 0.026 & 0.922 & 0.718 \\
\bottomrule
\end{tabular}
\caption{\textbf{TPR at stricter operating points.} Ungated Mahalanobis screen and ungated log-ratio $\rho$ (ablation statistic, not the frozen rule) on the overlapping split at 0.5\% and 0.1\% target FPR.}
\label{tab:tail}
\end{table}
\section{Full Detector List}
\label{sec:supp:detectors}

\Cref{tab:detectors} lists every baseline detector reported in this paper and how it is adapted to a GUI agent: TeCo, IBD-PSC, DeDe, STRIP and SCALE-UP adapt published methods, the Mahalanobis screen and the one-class SVM (OC-SVM) apply standard one-class detectors to $h_0$, and the others are generic uncertainty, occlusion and hidden-state scores.
Every detector is thresholded by split-conformal calibration at the target FPR (\cref{sec:supp:calib}), but on different clean sets: $B_1\cup B_2$ for the thirteen baselines of \Cref{tab:strict}, the calibration split of the earlier single split for the baseline rows of \Cref{tab:main,tab:backbone}, and 1000 clean calibration steps of each victim for the per-victim baselines of \Cref{tab:extra_baselines}.
Observed FPRs are therefore comparable only within a protocol.

\section{Calibration-size robustness}
\label{sec:supp:calibsize}

\begin{figure}[!htb]
\centering
\includegraphics[width=\linewidth]{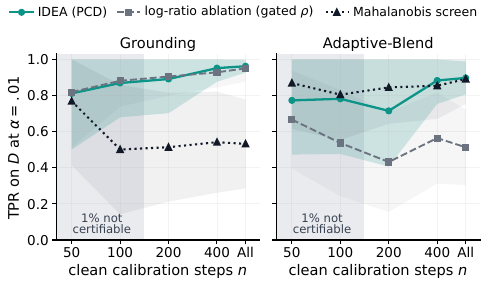}
\caption{\textbf{IDEA needs about 400 clean calibration steps.} TPR on followed steps (episode-disjoint split; mean and sd band over 20 seeds) with $n$ calibration steps; ``All'' is the whole pool. Shaded: only the attainable level ($1/26$, $1/51$) is certified.}
\label{fig:calib_size}
\end{figure}

\Cref{tab:calibsize} gives the calibration-size sweep for IDEA, its log-ratio ablation and the Mahalanobis screen on Grounding and Adaptive-Blend, from which \Cref{fig:calib_size} is drawn; \Cref{tab:sens} in the main text is an excerpt.
We subsample $n/2$ steps from each of $B_1$ and $B_2$ of the episode-disjoint split.
The fit split $F$ and the whitening stay fixed, so $n$ counts calibration steps only; the gate threshold and the clean law of \cref{eq:law} are refit on the subsampled $B_1$ at every size.
The whole pool (528--688 steps, about 600) reproduces \Cref{tab:strict}.
On Grounding, IDEA reaches 0.891 at $n=200$, 0.952 at $n=400$ and 0.962 on the whole pool, against 0.514, 0.542 and 0.532 for the Mahalanobis screen; the seed spread is large at $n=200$ (sd 0.189) and falls to 0.076 and 0.036.
The log-ratio is slightly ahead of IDEA up to $n=200$ (0.905 vs.\ 0.891), where the four coefficients of the clean law rest on 100 steps.
On Adaptive-Blend, the screen is ahead up to $n=200$ (0.845 vs.\ 0.715, IDEA seed sd 0.309); IDEA reaches 0.883 at $n=400$ and 0.898 on the whole pool, within seed spread of the screen (0.854 and 0.891), and the log-ratio stays at 0.43--0.67 at every size.
IDEA's curve on this family is not monotone: 0.782 at $n=100$ and 0.715 at $n=200$. The $n=100$ entry is measured at a looser level, since 50 threshold scores certify only $1/51\approx 2\%$ whereas 100 certify 1\%, and the seed sd is 0.31 at both sizes (a standard error of about 0.07 over 20 seeds), so the drop of 0.067 is within seed noise.
Below $n=200$, the score split cannot certify 1\% and the effective level rises (\eg, at $n=50$, FPR$_T$ is 0.020--0.038).

\begin{table}[!htb]
\centering
\scriptsize
\setlength{\tabcolsep}{4pt}
\begin{tabular}{llccc}
\toprule
Family & $n$ & IDEA & Log-ratio (abl.) & Screen \\
\midrule
Grounding & 50 & 0.812/0.027 & 0.819/0.027 & 0.769/0.038 \\
 & 100 & 0.869/0.012 & 0.882/0.014 & 0.501/0.011 \\
 & 200 & 0.891/0.008 & 0.905/0.007 & 0.514/0.009 \\
 & 400 & 0.952/0.008 & 0.929/0.008 & 0.542/0.010 \\
 & All & 0.962/0.008 & 0.950/0.008 & 0.532/0.009 \\
\midrule
Adaptive-Blend & 50 & 0.773/0.023 & 0.667/0.020 & 0.869/0.032 \\
 & 100 & 0.782/0.008 & 0.537/0.008 & 0.805/0.010 \\
 & 200 & 0.715/0.006 & 0.432/0.006 & 0.845/0.009 \\
 & 400 & 0.883/0.009 & 0.564/0.010 & 0.854/0.008 \\
 & All & 0.898/0.007 & 0.513/0.010 & 0.891/0.009 \\
\bottomrule
\end{tabular}
\caption{\textbf{Calibration-size sweep on the episode-disjoint split.} TPR on followed steps / pooled FPR$_T$, mean over 20 seeds, at the effective level $\max(0.01, 1/(n_{\mathrm{thr}}{+}1))$.}
\label{tab:calibsize}
\end{table}

\section{Localization Data}
\label{sec:supp:loc}

\Cref{tab:loc} provides the per-family localization metrics summarized in \cref{sec:loc}.
On the strict episode-disjoint split, the whitener is refit on $F$ for each of
 the 20 seeds.
The region $R^\star$ is the proposal that maximizes $E_j$,
 and we compute the rates over correct alarms of the frozen rule, defined as followed triggered steps with $S>T_\alpha$.
We report the mean over seeds; the seed standard deviation is at most 0.003 for
 every entry at $\alpha=0.01$.
The hit rate is the share of these alarms whose top proposal
 $R^\star$ has its center inside the trigger bounding box; we measure the IoU$>0.5$
 rate and the mean IoU (mIoU) against the same bounding box.
Because $R^\star$
 depends only on the audit, a change in the statistic only changes which steps count as alarms.
The seed standard deviation is at most 0.003 at $\alpha=0.01$ (0.006 for the Grounding bounding-box view, which was scored on the log-ratio's alarms and not rerun), and Popup never reaches IoU$>0.5$ because its banner is thinner than one grid cell.
The overlapping-split
 columns originate from an earlier single-split run (one seed, followed steps alarmed by the gated log-ratio), the
 only run that also scored the earlier subspace log-ratio variant of \Cref{tab:main}, which serves as
 the attribution baseline.

\paragraph{Readout of the qualitative examples.}
\Cref{tab:qual}
 lists the quantities computed by the audit for the five alarms shown in \Cref{fig:qual}: the presence $n_0$, the gate distance, the evidence and intent scales with their respective residuals to the clean law, the score
 against its threshold, and the IoU of $R^\star$ with the trigger box.

\begin{table}[!htb]
\centering
\footnotesize
\setlength{\tabcolsep}{3pt}
\renewcommand{\arraystretch}{0.95}
\resizebox{\columnwidth}{!}{%
\begin{tabular}{@{}lcccccccc@{}}
\toprule
Example & $n_0$ & $s_0\,/\,T_1$ & $E_{\max}$ & $S_{\mathrm{int}}$ & $u$ & $v$ & $\rho_{\mathrm{pc}}\,/\,T_\alpha$ & IoU \\
\midrule
(a) Patch & 293.5 & 150.6\,/\,50.6 & 155.1 & 94.9 & 4.40 & $+$5.54 & 7.07\,/\,0.64 & 0.54 \\
(b) Popup & 202.7 & 98.4\,/\,52.2 & 101.3 & 44.5 & 4.10 & $+$9.46 & 10.32\,/\,0.27 & 0.16 \\
(c) Grounding & 103.1 & 53.6\,/\,51.6 & 47.7 & 61.4$^\ast$ & 1.35 & $+$3.63 & 3.87\,/\,1.15 & 0.29 \\
(d) TypeText & 292.3 & 148.4\,/\,50.6 & 145.6 & 59.0$^\ast$ & 3.88 & $+$8.61 & 9.45\,/\,0.50 & 0.40 \\
(e) Adaptive-Blend & 142.9 & 77.4\,/\,51.4 & 90.5 & 141.8 & 5.00 & $-$1.73 & 5.00\,/\,0.84 & 0.00 \\
\bottomrule
\end{tabular}%
}
\caption{\textbf{Readout of the five alarms in \Cref{fig:qual}.} Gate distance $s_0$ against $T_1$, the two scales, their residuals to the clean law, and the score against $T_\alpha$. $^\ast$Fallback median.}
\label{tab:qual}
\vspace{-2mm}
\end{table}